\pdfoutput=1

\documentclass[a4paper, 11pt, abstracton, titlepage, oneside]{scrartcl}
\makeatletter

\usepackage[T1]{fontenc}
\usepackage[utf8]{inputenc}
\usepackage[onehalfspacing]{setspace}
\usepackage[headsepline]{scrlayer-scrpage}
\clearscrheadfoot 
\usepackage{layout}
\usepackage{geometry}
\usepackage{adjustbox}
\usepackage{authblk}
\usepackage[titletoc,title]{appendix}

\usepackage[hyphens]{url}
\usepackage{color}

\usepackage{amsmath,amssymb,amsfonts,amsthm, mathtools, bm}
\DeclareMathOperator*{\argmax}{arg\,max}

\usepackage{thmtools, thm-restate}
\usepackage{nicefrac}
\usepackage{array}

\usepackage[normal]{threeparttable}
\usepackage{threeparttablex}
\usepackage{tabularx}
\usepackage{longtable}
\usepackage{booktabs}
\usepackage{colortbl}
\usepackage{multirow}
\usepackage{makecell}

\usepackage[acronym]{glossaries}

\usepackage{subcaption}
\usepackage{floatrow}
\usepackage{graphicx}
\usepackage{placeins}

\usepackage{tikz}
\usetikzlibrary{arrows.meta}
\usetikzlibrary{math}
\usetikzlibrary{shapes}
\usepackage{pgfplots}
\usepgfplotslibrary{groupplots}
\usepgfplotslibrary{dateplot}
\usepackage{tikzscale}
\usetikzlibrary{patterns}

\usepackage{enumerate}
\usepackage{multicol}
\usepackage[author=Patrick]{pdfcomment}
\usepackage{xparse}
\usepackage{twoopt}
\usepackage{etoolbox}

\usepackage{eurosym}
\usepackage{ellipsis}
\usepackage{natbib}
\usepackage{paralist}
\usepackage{ulem}

\usepackage{enumitem} 
\usepackage{algpseudocode}
\usepackage{algorithm}
\usepackage{caption}

\AtBeginDocument{
	\newgeometry{
		textheight=9in,
		textwidth=6.5in,
		top=1in,
		headheight=14pt,
		headsep=25pt,
		footskip=30pt
	}
}
\newsavebox{\abstractbox}
\renewenvironment{abstract}
{\begin{lrbox}{0}\begin{minipage}{\textwidth}
			\begin{center}\normalfont\sectfont\abstractname\end{center}\quotation}
		{\endquotation\end{minipage}\end{lrbox}%
	\global\setbox\abstractbox=\box0 }

\makeatletter
\expandafter\patchcmd\csname\string\maketitle\endcsname
{\vskip\z@\@plus3fill}
{\vskip\z@\@plus2fill\box\abstractbox\vskip\z@\@plus1fill}
{}{}
\makeatother

\addtokomafont{captionlabel}{\footnotesize\bfseries} 
\addtokomafont{caption}{\footnotesize\bfseries} 

\bibpunct[, ]{(}{)}{,}{a}{}{,}%
\counterwithin*{equation}{section}

\newenvironment{myfont}{\fontfamily{ccr}\selectfont}{\par}
\DeclareTextFontCommand{\textmyfont}{\myfont}

\AtBeginDocument{%
	\abovedisplayskip=7.5pt plus 0pt minus 0pt
	\abovedisplayshortskip=7.5pt plus 0pt minus 0pt
	\belowdisplayskip=7.5pt plus 0pt minus 0pt
	\belowdisplayshortskip=7.5pt plus 0pt minus 0pt
}

\newcolumntype{L}[1]{>{\raggedright\let\newline\\\arraybackslash\hspace{0pt}}p{#1}}
\newcolumntype{C}[1]{>{\centering\let\newline\\\arraybackslash\hspace{0pt}}p{#1}}
\newcolumntype{R}[1]{>{\raggedleft\let\newline\\\arraybackslash\hspace{0pt}}p{#1}}
\renewcommand{\emph}[1]{\textit{#1}}

\usepackage{comment}

\newtheorem{result}{Result}

\begin{document}
\emergencystretch 3em
\title{\large Optimization-Augmented Machine Learning for Vehicle Operations in Emergency Medical Services}

\author{\normalsize Maximiliane Rautenstrauß\textsuperscript{1}}
\author{\normalsize Maximilian Schiffer\textsuperscript{1,2}}
\affil{\small 
        \textsuperscript{1}School of Management, Technical University of Munich, Germany
        
        \scriptsize maximiliane.rautenstrauss@tum.de
	
	\small
	\textsuperscript{2}Munich Data Science Institute, Technical University of Munich, Germany
        
        \scriptsize schiffer@tum.de}

\date{}

\lehead{\pagemark}
\rohead{\pagemark}

\begin{abstract}
\begin{singlespace}
{\small\noindent Minimizing response times to meet legal requirements and serve patients in a timely manner is crucial for Emergency Medical Service (EMS) systems. Achieving this goal necessitates optimizing operational decision-making to efficiently manage ambulances. Against this background, we study a centrally controlled EMS system for which we learn an online ambulance dispatching and redeployment policy that aims at minimizing the mean response time of ambulances within the system by dispatching an ambulance upon receiving an emergency call and redeploying it to a waiting location upon the completion of its service. We propose a novel combinatorial optimization-augmented machine learning pipeline that allows to learn efficient policies for ambulance dispatching and redeployment. In this context, we further show how to solve the underlying full-information problem to generate training data and propose an augmentation scheme that improves our pipeline's generalization performance by mitigating a possible distribution mismatch with respect to the considered state space. Compared to existing methods that rely on augmentation during training, our approach offers substantial runtime savings of up to 87.9\% while yielding competitive performance. To evaluate the performance of our pipeline against current industry practices, we conduct a numerical case study on the example of San Francisco’s 911 call data. Results show that the learned policies outperform the online benchmarks across various resource and demand scenarios, yielding a reduction in mean response time of up to 30\%. \\
\smallskip}
\end{singlespace}
\end{abstract}

\maketitle
\section{Introduction}
Minimizing response times to meet legal requirements and serve patients in a timely manner is crucial for Emergency Medical Service (EMS) systems.  Numerous studies show a significant increase in patients' survival rates when reducing response times. For example, a Canadian study shows a steep survival chance reduction within the first five minutes following cardiac arrest. With each additional minute of delay to defibrillation, the chance of survival further decreases by 23\% \citep{de2003optimal}. 
Despite its importance, reducing response times remains challenging for EMS systems. One contributing factor is the widespread shortage of paramedics  \citep[see, e.g., ][]{quaile2015tackling, whitfield2020critical}. Increasing ambulance demand, intense work pressure, and long shifts have led to a significant number of paramedics leaving their profession \citep{quaile2015tackling}. Consequently, as demand for ambulance services faces an upward trend, maintaining or even reducing response times despite limited personnel necessitates optimizing operational decision-making to efficiently manage scarce ambulance resources.

In this context, recent work has focused on developing models for dispatching ambulances and for determining optimal waiting positions for idle ambulances that enable a fast response to future requests. To allow an application in real-time settings, these models must be computationally efficient. In addition, they need to capture the stochastic nature of EMS processes. In this context, the uncertainty associated with future requests, encompassing the incidents' locations, service times, and patients' drop-off locations, poses a challenge as this information remains unknown at the time of dispatching. 

In practice, dispatchers mainly apply heuristics based on fixed decision rules, e.g., always dispatching the closest available ambulance to an incident and redeploying the vehicle to a fixed station after service completion. Such approaches allow for fast application; however, they do not take advantage of anticipatory decision-making, e.g., by taking into account potential future emergency calls. This may result in regions being left with insufficient ambulance coverage, resulting in extended response times. 

In academia, such decision-making problems are often modeled as a Markov Decision Process (MDP), which paves the way for applying approximate dynamic programming (ADP) or (deep) reinforcement learning techniques to obtain tangible decision-making policies. These approaches benefit from processing contextual information when making decisions but may struggle to capture combinatorial elements of the underlying decision-making problem.

\paragraph{Contribution.}
Against this background, we propose a novel combinatorial optimization (CO)-augmented machine learning (ML) approach to encode an effective policy for ambulance dispatching and redeployment decisions that combines the advantages of both domains. 
Specifically, our methodological contribution is threefold: first, we propose a CO-augmented ML pipeline for ambulance dispatching and redeployment in real-time. This pipeline combines an ML-layer that allows to process context and capture uncertain dynamics with a CO-layer that allows to incorporate the underlying decision-making problem's combinatorial structure. Second, we show how to parameterize the statistical model in the ML-layer in an end-to-end fashion by leveraging a structured learning approach. In this context, we propose a novel augmentation scheme that improves the pipeline's generalization performance by modifying the training data apriori to the training process. Third, we show how to efficiently solve our decision-making problem's full information counterpart to create training data for the structured learning approach.
 
To evaluate the performance of the proposed pipeline, we benchmark the learned online policies against current industry practices based on real data from San Francisco.
Results show that the learned policies can outperform the online benchmarks across all scenarios, varying both available resources and emergency call volumes. Classical approaches such as ADP yield improvements of approximately 7\% to 10\% over current industry practices in comparable setups \citep{schmid2012solving, nasrollahzadeh2018real} while our learned policies enable mean response time reductions of up to 30\%. Enhancing the training set by states generated from suboptimal dispatching and redeployment policies can further improve the policies' performances. The learned policies benefit from anticipatory decision-making, resulting in shorter driving distances and reduced response times. We further show that our augmentation scheme yields significant runtime savings of up to 87.9\% compared to existing schemes that rely on augmentation during training while preserving a competitive performance.

\paragraph{Organization.} The remainder of this paper is structured as follows.  We outline related work in Section \ref{sec:literature}. Section \ref{sec:problem_description} introduces our problem setting, while Section \ref{sec:methodology} develops the methodology needed to establish the CO-augmented ML pipeline. In Section \ref{sec:castestudy}, we introduce a numerical case study for San Francisco’s 911 call data and discuss numerical results in Section \ref{sec:casestudy_results}. Section \ref{sec:conclusion} concludes this work.

\section{State of the Art}\label{sec:literature}

This research connects two fields of literature: It addresses optimizing EMS operations from an application viewpoint and aligns with CO-augmented ML methodologically. In this section, we discuss related work from both domains.

\paragraph{Optimization of Emergency Service Operations.}
Optimizing operational decisions in EMS systems has gained a lot of attention in recent years. Recent studies mainly focus on \textit{dispatching}, \textit{redeploying}, and \textit{relocating} ambulances to minimize response times. 
Both redeployment and relocation decisions focus on finding optimal waiting positions for ambulances from where they can respond to future requests. We distinguish both terms as follows: While redeployment allocates an ambulance directly after serving a request to a waiting position, relocation focuses on moving idle ambulances from one waiting position to another. Although many studies show the effectiveness of improving redeployment and relocation decisions, in practice, ambulances are often allocated to fixed bases, which they always return to after serving a request \citep{schmid2012solving}.  In this context, static models such as the \textit{maximum covering location problem} (MCLP) \citep{church1974maximal} or the \textit{maximum expected covering location problem} (MEXCLP) \citep{daskin1983maximum}, and related extensions are often applied at the tactical level for allocating ambulances to bases. 
Many works leverage such static models, e.g., by extending and applying them in a rolling horizon fashion, to develop dynamic approaches for real-time relocation and redeployment applications, see, e.g., \cite{gendreau2001dynamic}, \cite{moeini2015location}, and \cite{jagtenberg2015efficient}. To reduce the computational effort caused by repetitive calculations, many studies focus on so-called compliance tables computed only once a priori. Compliance tables define the required allocation of ambulances to waiting positions for all possible system states.
 In this context, studies focus on evaluating \citep[see, e.g., ][]{alanis2013markov} and optimizing \citep[see, e.g., ][]{sudtachat2016nested} compliance tables.
 More recent works, such as \cite{ji2019deep} and \cite{elfahim2022deep}, apply deep reinforcement learning to optimize the redeployment of ambulances. A complementary stream of literature focuses on the optimization of dispatching policies. While few models focus solely on dispatching decisions, e.g., \cite{jagtenberg2017optimal} and \cite{hua2022optimal}, most models aim at jointly optimizing dispatching and repositioning decisions. For example, \cite{andersson2007decision} use a region's \textit{preparedness} measure quantifying the system's ability to respond to future requests in a heuristic to dispatch and relocate ambulances. \cite{naoum2013stochastic} apply two-stage stochastic programming to relocate ambulances at the first stage and make dispatching decisions at the second stage. Many studies model the problem as an MDP \citep[see, e.g., ][]{schmid2012solving}, optimizing redeployment and dispatching decisions by applying ADP. \cite{nasrollahzadeh2018real} further extend this ADP approach to include relocation decisions alongside dispatching and redeployment. We refer to \cite{brotcorne2003ambulance}, \cite{aringhieri2017emergency}, and \cite{belanger2019recent} for extensive literature reviews on dispatching, redeployment, and relocation approaches for EMS systems.
 
We note that most studies focus on deriving dispatching, redeployment, and relocation policies for online settings to enable an application in practice. Solely \cite{jagtenberg2017benchmarking} present a benchmark model to derive optimal dispatching decisions for offline settings where future requests are presumed to be known. However, a model for optimizing both dispatching and redeployment decisions assuming full information availability has not yet been developed. While our work focuses on a new pipeline for online decision-making, we additionally fill this gap by presenting a methodology that allows to efficiently compute optimal dispatching and redeployment decisions in a full information setting. This yields a valuable full-information benchmark, enabling the quantification of an online algorithm's performance.
 
 \paragraph{Combinatorial Optimization augmented Machine Learning.}\label{sec:literature_CO_enriched_ML-pipelines}
 Recently, contextual optimization has gained increasing attention for addressing multi-stage decision-making problems in which decision-makers make anticipative decisions in real-time, e.g., for optimizing vehicle routing, resource scheduling, and inventory management decisions. In such settings, parameters, e.g., future travel times or demands, are often unknown. By combining ML and CO, ML components can be leveraged to process contextual information, e.g., by predicting unknown parameters, for solving the CO problem. In this context, a commonly applied paradigm is \textit{predict-then-optimize}, which, first, predicts the model's unknown parameters, and subsequently applies CO to the parameterized model. In the realm of EMS systems, \cite{mukhopadhyay2017prioritized} apply this paradigm by first predicting emergency incidents and, subsequently, solving a CO problem to determine the optimal ambulance waiting locations and ambulance allocation. \cite{boutilier2020ambulance} make ambulance location and routing decisions by first predicting ambulance demand and travel times and second, solving a two-stage optimization model. \cite{takedomi2022facility} use this paradigm to address the optimization of emergency road service facility locations. As a basis, a predictive model estimates the vehicles' preparation and travel times based on contextual information. These predictions are then used by a CO model to select the optimal subset of candidate locations for positioning service facilities. 
Although widely used, this paradigm typically focuses solely on minimizing the prediction error when training the ML-layer. Consequently, it has a notable limitation: training the ML-layer ignores the predictions' influence on the subsequent optimization tasks. Specifically, it falls short in considering the decision error induced by the prediction. For this reason, \textit{smart predict-then-optimize} pipelines train the ML predictor with a loss function quantifying the decision error \citep[see, e.g.,][] {elmachtoub2022smart}. However, in many real-world scenarios, knowing the true cost vector of the optimization problem, which is essential for this approach, may be unknown or computationally expensive to determine. Recently, {\textit{end-to-end learning} of CO-augmented ML pipelines has proven to be a promising approach for making anticipative online decisions in the context of multi-stage decision-making problems, addressing the aforementioned drawbacks. In these pipelines, an ML model is trained to parameterize a CO problem. The solution to the parameterized CO problem provides a feasible solution to the original problem. The aim is to enable the ML model to generate parameterizations for the CO model that minimize the deviation between the predicted and the optimal solution in an imitation learning fashion. Applying a perturbed Fenchel Young loss to quantify the non-optimality between the predicted solution and its optimal solution enables the minimization of the loss function via stochastic gradient descent \citep{berthet2020learning}.
This paradigm has recently been applied to various applications, including the stochastic vehicle scheduling problem
 \citep{parmentier2022learning}, the single-machine scheduling problem \citep{dalle2022learning, parmentier2023structured}, the two-stage stochastic minimum spanning tree problem \citep{dalle2022learning}, and dynamic vehicle routing problems in the context of last-mile distribution and autonomous mobility-on-demand systems \citep{baty2023combinatorial,jungel2023learning,greif2024combinatorial}.  

 While the latter works give hope that CO-augmented ML pipelines may allow to learn efficient policies for ambulance dispatching and redeployment, a problem-specific pipeline has not been developed so far. Developing such a pipeline requires tailoring novel optimization problems and identifying a suitable statistical model for the ML-layer. Beyond these technical challenges, there exists one central difference between the ambulance dispatching and redeployment problem studied in this work and existing works on applying CO-augmented ML pipelines to transportation problems: so far, all existing works focused on applications for which big data was available. In the context of ambulance operations, the daily patterns exhibit much fewer data points, such that our work is the first to apply such a pipeline to a small data context.

\section{Problem Setting}\label{sec:problem_description}
We study an online dispatching and redeployment policy for ambulances operating in a centrally controlled EMS system. The EMS operates in a specified region where emergency requests arise at spatially distributed locations. Receiving a request results in a prompt dispatch of an ambulance. In the case that no ambulance is idle at the time the request enters the system, we assume that it joins a queue to which ambulances are assigned using a first-in-first-out (FIFO) principle. We account for emergencies requiring the response of multiple ambulances by creating the corresponding number of requests with a marginal time delay, ensuring a sequential arrival. After being dispatched, the ambulance drives to the incident scene. In line with \cite{schmid2012solving, maxwell2014bound}, and \cite{jagtenberg2017optimal}, we neglect the turnout time of ambulances, i.e., the time it takes the assigned ambulances to leave their current waiting position after being dispatched to a request. After arrival, the ambulance serves the request. Its service time includes the patient's treatment time at the incident's location and, if necessary, the transport and drop-off at the hospital. After providing service, the ambulance must be directly sent to a queued request or redeployed to a waiting location, which we model as a redeployment request. We aim at learning an online dispatching and redeployment policy that minimizes the mean response time within the system.
\paragraph{Notation.} Let $\mathcal{M}$ be the fleet of ambulances operated by the EMS system, and let $\mathcal{L}$ be the set of locations within the designated region. We denote the set of incoming requests by $\mathcal{R} = \mathcal{\tilde{R}} \cup \mathcal{\hat{R}}$ consisting of emergency requests $\mathcal{\tilde{R}}$ and redeployment requests $\mathcal{\hat{R}}$. While this distinction may appear unnecessary at first, it becomes valuable for defining performance measures and formulating constraints in the following. We describe each request $r \in \mathcal{R}$ by a quadruple $r=\left(o_r, e_r, d_r, s_r\right)$ where $o_r \in \mathcal{L}$, $e_r \in \mathbb{R}_{\geq 0}$, $d_r \in \mathcal{L}$ and $s_r \in \mathbb{R}_{\geq 0}$ denote the incident location, the time at which it enters the system, the patient's drop-off location, and the required service time, respectively. If a patient is transported to a hospital, we set the drop-off location $d_r$ to the hospital's location. Otherwise, the drop-off location coincides with the incident's location, i.e., $o_r=d_r$. After completing its service, an ambulance can be redeployed to any waiting location $\mathcal{L}^R \subseteq \mathcal{L}$. For redeployment requests, we set $o_r$ and $d_r$ to the waiting location $l \in \mathcal{L}^R$ and set the service time $s_r$ to $0$ as no treatment takes place. We denote the driving time between locations $l, l' \in \mathcal{L}$ by $\tau\left(l, l'\right)$. The slack between two successive requests $(r, r')$ is 
\begin{equation}
    \delta_{rr'}=e_{r'} - e_{r} - s_{r}.
\end{equation}
Given a pair of successivly served requests $(r, r')$, let the response time $\upsilon_{r'}$ of request $r'$ be  
\begin{equation}
    \upsilon_{r'} = \max(c_{r}, e_{r'})+ \tau(d_{r}, o_{r'}),
\end{equation}
where the completion time of $r'$ can be derived by 
\begin{equation}
    c_{r'} = \max(c_{r}, e_{r'}) + \tau(d_{r}, o_{r'}) + s_{r'}.
\end{equation}
With this notation, we formalize our decision-making problem as an MDP. Let $\mathcal{T} = \{0,1,...,T\}$ denote our time horizon in which the system evolves event-based at discrete time steps $t \in \mathcal{T}$, based on the current state $x_t \in \mathcal{X}_t$ and chosen actions $y_t \in \mathcal{Y}(x_t)$ which we define in the following. 
\paragraph{System State.} We denote the system state at time $t$ by $x_t = (\mathcal{R}_t, (l_{m},\mathcal{S}_{m,t})_{m \in \mathcal{M}})$ which comprises, for each ambulance $m \in \mathcal{M}$, the initial location $l_{m}$ and a sequence of assigned requests ${\mathcal{S}_{m,t}=\{r, ..., r'\}}$ up to timestamp $t$. We note that there is no need to include the location of an ambulance at timestamp $t$ in $x_t$ as it can be easily extracted from $\mathcal{S}_{m,t}$. $\mathcal{R}_t$ denotes a set of requests arriving at time $t$. 
\paragraph{Transition Dynamics.} The system evolves on an event basis. Specifically, upon the arrival of a request batch $\mathcal{R}_t$ at time $t$, the dispatcher immediately takes a dispatching or redeployment decision $y_t \in \mathcal{Y}_t$. While emergency requests arrive individually, redeployment requests arrive in batches. Each batch of redeployment requests includes all possible redeployment options. We assume that emergency and redeployment requests do not occur simultaneously, ensuring that $\mathcal{R}_t$ contains either an emergency request or a batch of redeployment requests. This assumption is justified by the fact that our system is not based on fixed time intervals but permits arbitrarily small time gaps between request arrivals. In the case of concurrent arrival, emergency calls always take precedence over redeployments. The transition function $f$ maps the state $x_t$ to its subsequent state $x_{t+1}$ taking into account the action $y_t$ taken at time $t$, and the incoming requests at time $t+1$ represented by $\mathcal{R}_{t+1}$: 
\begin{align}\label{eq:transition}
x_{t+1} &= f(x_t, y_t, \mathcal{R}_{t+1}).
\end{align}
\paragraph{Action Space.} A feasible decision $y_{t} \in  \mathcal{Y}(x_{t})$ consists of assigning an ambulance $m$ to a request $r \in R_t$ and updating the sequence of assigned requests $\mathcal{S}_{m,t}$. This assignment represents either a dispatch to an emergency request or a redeployment to a waiting location. We note that direct dispatches between emergency requests are permitted only if the succeeding request $r'$ arrives before the preceding request $r$ is completed (i.e., $c_{r} \geq e_{r'}$) to prevent ambulances from idling at unspecified locations. 

\paragraph{Objective.} We aim at learning a policy $\pi: \mathcal{X} \rightarrow \mathcal{Y}$ that maps a problem instance $x \in \mathcal{X}$ to a feasible solution $y \in \mathcal{Y}(x)$ that minimizes the mean response time $\Upsilon_\mathcal{T}$ over all requests that arrive in our time horizon $\mathcal{T}$:
\begin{equation}
    \min_{\pi} \mathbb{E} ( \Upsilon_\mathcal{T} | \pi),
\end{equation}
where we define the mean response time $\Upsilon_t$ for $x_t$ by
\begin{equation}
    \Upsilon_t = \frac{1}{|\mathcal{\tilde{R}}_t|} \sum_{r \in \mathcal{\tilde{R}}_t} \upsilon_{r}.
\end{equation}

\section{Methodology}\label{sec:methodology}
{\color{black} In the following, we develop a CO-augmented ML pipeline as shown in Figure~\ref{fig:pipeline} that allows to compute effective policies for the ambulance dispatching and redeployment problem as introduced in Section~\ref{sec:problem_description}. 
Such a pipeline combines a ML-layer with an CO-layer, which allows to combine the strengths of statistica modeling with the strengths of discrete optimization: we will use the statistical model $\varphi_w$ in the ML-layer, which receives a system state $x_t$ as input, to predict a parameterization $\theta$. We will then use this parameterization $\theta$ as input for the CO-layer, specifically to encode an instance for the respective optimization problem, which we then solve to derive a feasible decision $y_t \in \mathcal{Y}(x_t)$. To learn the parameterization $w$ of our statistical model $\varphi_w$, we leverage a structured learning approach, i.e., imitation learning in a combinatorial space: we aim to parameterize $\varphi_w$ such that the distance between a reference policy $\pi'$ and the policy $\hat{\pi}$ encoded in our pipeline becomes minimal.
\begin{figure}[h!]
  \centering
  \includegraphics[width=.75\linewidth]{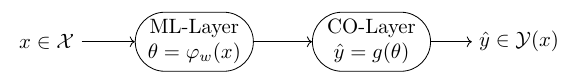}
\caption{Optimization-Augmented Machine Learning Pipeline Architecture}
\label{fig:pipeline}
\end{figure}

Developing such a pipeline requires the following methodological steps: in a first step, we need to derive training data by computing a reasonable reference policy $\pi'$. In a second step, we need to specify the pipeline's architecture, i.e., its ML-layer and its CO-layer. In a final step, we need to specify the (structured) learning paradigm that allows to parameterize the statistical model~$\varphi_w$. In the remainder of this section, we detail these three steps for our ambulance dispatching and redeployment problem.}

\subsection{Offline Decision-Making: Generating Training Data}\label{sec:methodology_generating_training_data}
{\color{black} To train a policy $\hat{\pi}$ based on a reference policy $\pi'$, we aim at deriving a set of $n$ training points $\{(x_i,y_i')\}_{i=1}^{n}$, which allows us to minimize the distance between the reference decisions $y_i'$ and the predicted decisions $\hat{y}_i$ derived by our pipeline, for the respective states $x_i$. As both policies $\pi'$ and $\hat{\pi}$ constitute a mapping between the state space $\mathcal{X}$ and the decision space $\mathcal{Y}$ as defined in Section~\ref{sec:problem_description}, minimizing the distance between decisions $\hat{y}_i$ and $y_i'$ also minimizes the distance between $\hat{\pi}$ and $\pi'$.

In the following, we detail how one can derive an anticipative set of training points $\mathcal{D}$ as it has recently been proposed by \cite{jungel2023learning} and \cite{baty2023combinatorial}. In Section 4.3, we will detail on how to enhance such a training set to improve the learning paradigms generalization performance. 

Building an anticipative training set requires two steps: first, solving the respective planning problem in a full-information, offline problem setting based on past data, and second, slicing the respective full-information solutions into training points $(x_i,y_i')$. In the following, we detail both steps.}

\paragraph{Solving the Offline Problem.}
{\color{black} We study the problem setting as described in Section~\ref{sec:problem_description} but assume full information about all requests arriving in $\mathcal{T} = \{0,...,T\}$ already in $t=0$, which allows us to solve our online problems' offline counterpart. To solve this problem efficiently, we derive a problem specific, acyclic graph structure, which allows to encode problem constraints in the graph itself, and subsequently formulate a MILP that allows to compute an optimal solution.}

To reduce the problem's complexity in the MILP, we construct an acyclic dispatching digraph $G_z = (\mathcal{V}, {\mathcal{A}})$ defined by a set of vertices $\mathcal{V}$ and arcs ${\mathcal{A}}$. The main rationale of this graph is that the vertices represent either initial ambulance locations at $t=0$, emergency requests, or possible redeployments, while each arc indicates a feasible dispatching decision. In such a graph, each path from source to sink constitutes a feasible solution for one vehicle. Accordingly, each set of $|\mathcal{M}|$ vertex-disjoint paths constitutes a feasible solution to our problem. Hence, this graph structure allows us to encode problem specific feasibility constraints in the time dimension, e.g., maximum driving times between locations or time-based restrictions, directly within the graph, which allows us to formulate an efficient MILP to compute an optimal solution.  

Formally, we construct this dispatching graph $G_z = (\mathcal{V}, {\mathcal{A}})$ as follows: the vertex set ${\mathcal{V}=\mathcal{V}^\mathcal{M}, \mathcal{V}^{\mathcal{\tilde{R}}}, \mathcal{V}^{\mathcal{\hat{R}}} \cup \{o, d\}}$ comprises ambulance vertices $\mathcal{V}^\mathcal{M}$, emergency vertices $\mathcal{V}^{\mathcal{\tilde{R}}}$, redeployment vertices $\mathcal{V}^{\mathcal{\hat{R}}}$, a dummy source vertex $o$ and a dummy sink vertex $d$. Then, we construct the arc set as follows:
\begin{enumerate}
\item We connect the dummy source vertex to every ambulance vertex. 
\item We connect every ambulance vertex to every emergency vertex representing possible dispatches. \item Every emergency vertex connects to its corresponding $|\mathcal{L}^R|$ redeployment vertices, modeling possible redeployment options. 
\item To allow subsequent dispatches from these waiting locations, we link every redeployment vertex $v_{r} \in \mathcal{V}^{\mathcal{\hat{R}}}$ to all emergency vertices $v_{r'} \in {\mathcal{V}}^{\mathcal{\tilde{R}}}$ for which $e_{r} < e_{r'}$. 
\item We introduce conditional arcs between successive requests vertices $(v_{r}, v_{r'} \in {\mathcal{V}}^{\mathcal{\tilde{R}}})$ if $e_{r} < e_{r'}$ to allow direct dispatches. These arcs can only be traversed if $c_r \ge e_{r'}$.
\item Finally, all redeployment and ambulance vertices are connected to the dummy sink vertex. 
\end{enumerate}
Figure~\ref{fig:offline_graph} shows an example of such a graph construction. 
\begin{figure}[h!]
\centering
  \includegraphics[width=.8\linewidth, trim={0 .3cm 0 .5cm}, clip]{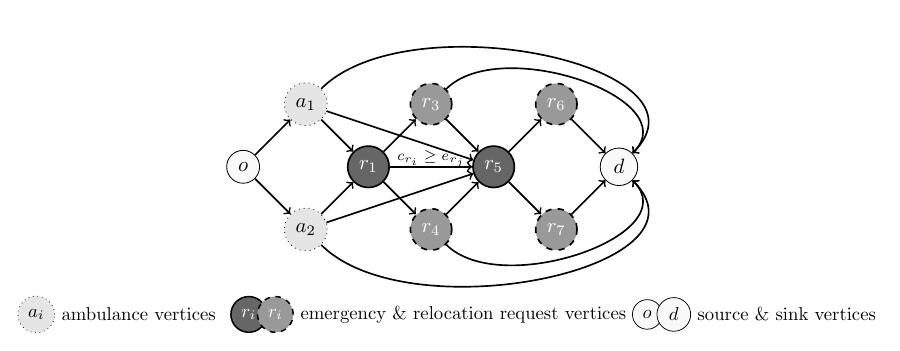}
  \caption{Offline Dispatching Digraph}
  \label{fig:offline_graph}
\end{figure}
Based on this graph, we solve the following mixed integer program (MIP) in a full-information setting, i.e., assuming that all future requests and request properties are known, to obtain optimal dispatch and redeployment decisions that minimize the mean response time. With a slight abuse of notation, we refer to the slack and driving time associated with an arc $(i,j) \in {\mathcal{A}}$ by $\delta_{ij}$ and $\tau_{ij}$, correspondingly. The variables $\alpha_{ij}$ and $\beta_{ij}$ denote the considered response times and delays, i.e., the elapsed time from an ambulance receiving a dispatch to the time it initiates driving towards this request. The binary variable $x_{ij}$ is set to 1 if an arc belongs to a path and to 0 otherwise. With this notation, our MIP is as follows.
\begin{equation}
    min \sum_{\substack{(i,j)\in {{\mathcal{A}}}:\\j\in {\mathcal{V}}^{\mathcal{\tilde{R}}}}} \alpha_{ij} \label{eq:MIP_objective}
\end{equation}
s.t.
\begin{align}
\alpha_{ij} &= \label{eq:calculation_response_times}
    \begin{cases}
        \left(\tau_{ij} + \beta_{ij}\right) x_{ij} & \forall (i, j) \in {{\mathcal{A}}}: j \in {\mathcal{V}}^{\mathcal{\tilde{R}}} \cup {\mathcal{V}}^{\mathcal{\hat{R}}}\\
        \mathrlap0\hphantom{\max{\left(\zeta_{ij} - \sum_{k \in {{\mathcal{V}}}: (k, i) \in {{\mathcal{A}}}}  \alpha_{ki}, 0\right)}} & \forall (i, j) \in {{\mathcal{A}}}: j \in {\mathcal{V}}^\mathcal{M} \cup \{d\} 
    \end{cases}\\
\beta_{ij} &= \label{eq:dispatch_delays}
    \begin{cases}
        \max{\left(-\delta_{ij} + \sum_{\substack{k \in {{\mathcal{V}}}:\\(k, i) \in {{\mathcal{A}}}}}  \alpha_{ki}, 0\right)} & \forall (i, j) \in {{\mathcal{A}}}: j \in {\mathcal{V}}^{\mathcal{\tilde{R}}} \cup {\mathcal{V}}^{\mathcal{\hat{R}}}\\
        0 & \forall (i, j) \in {{\mathcal{A}}}: j \in {\mathcal{V}}^\mathcal{M} \cup \{d\} 
    \end{cases}
\end{align}
\begin{align}
\sum_{i\in{{\mathcal{V}}}} x_{ij} &= 1 &&\forall j \in {\mathcal{V}}^{\mathcal{\tilde{R}}} \cup {\mathcal{V}}^{\mathcal{M}}: (i, j) \in {{\mathcal{A}}} \label{eq:request_assigned_to_max_1_path} \\
x_{ij} &= 0 &&\forall (i, j) \in {{\mathcal{A}}}: i, j \in {\mathcal{V}}^{\mathcal{\tilde{R}}} \land  \beta_{ij} = 0 \label{eq:direct dispatches} \\
\sum_{\substack{j \in {{\mathcal{V}}}:\\(j, i) \in {{\mathcal{A}}}}} x_{ji} &= \sum_{\substack{j \in {{\mathcal{V}}}:\\(i, j) \in {{\mathcal{A}}}}} x_{ij} &&\forall i \in {{\mathcal{V}}} \setminus \{s,d\} \label{eq:flow_conservation} \\
x_{ij} &\in \{0,1\} &&\forall (i, j) \in {{\mathcal{A}}} \label{eq:variable_domain_x} \\
\beta_{ij}, \alpha_{ij} &\in \mathbb{R}_{\leq 0}&&\forall (i, j) \in {{\mathcal{A}}} \label{eq:variable_domain_alpha}
\end{align}
Objective (\ref{eq:MIP_objective}) minimizes the sum of response times defined in (\ref{eq:calculation_response_times}). Constraint (\ref{eq:dispatch_delays}) calculates the delays. Constraint (\ref{eq:request_assigned_to_max_1_path}) ensures that a request is served by exactly one ambulance, and assigns each ambulance to exactly one path. We note that connecting ambulance vertices to the sink vertex allows an ambulance to remain unused. Constraint (\ref{eq:direct dispatches}) enables direct dispatches if and only if a request is queued. Constraints (\ref{eq:variable_domain_x}) and (\ref{eq:variable_domain_alpha}) define the variable domains.

We transform the model into a MILP by linearizing the maximization terms in (\ref{eq:dispatch_delays}) and the quadratic constraints in (\ref{eq:calculation_response_times}). We apply McCormick inequalities \citep{mccormick1976computability} for linearization of (\ref{eq:calculation_response_times}) as follows:
\begin{align}
    w_{ij} &= \beta_{ij}x_{ij}\\
    w_{ij} &<= \beta_{ij}^Lx_{ij} + \beta_{ij}x_{ij}^U - \beta_{ij}^Lx_{ij}^U\\
    w_{ij} &<= \beta_{ij}^Ux_{ij} + \beta_{ij}x_{ij}^L - \beta_{ij}^Ux_{ij}^L\\
    w_{ij} &>= \beta_{ij}^Lx_{ij} + \beta_{ij}x_{ij}^L - \beta_{ij}^Lx_{ij}^L\\
    w_{ij} &>= \beta_{ij}^Ux_{ij} + \beta_{ij}x_{ij}^U - \beta_{ij}^Ux_{ij}^U
\end{align}
where
\begin{align}
    \beta_{ij}^L &<= \beta_{ij} <= \beta_{ij}^U\\
    x_{ij}^L &<= x_{ij} <= x_{ij}^U
\end{align} 
are the upper and lower bounds of $\beta_{ij}$ and $x_{ij}$, correspondingly. The linearization of the maximization term in (\ref{eq:dispatch_delays}) is trivial.

{\color{black}The introduced model allows us to incorporate problem specific requirements, e.g., maximum driving time constraints, by transforming the underlying graph. However, the model may become infeasible as we enforce all requests to be served. For this reason, we present an alternative modeling approach in Appendix \ref{sec:appendix_solution_model_with_soft_constraints} which maximizes the number of served requests while enabling requests to be dropped. We compare the models' runtimes in Appendix \ref{sec:appendix_solution_model_with_soft_constraints}. As the assumptions made in the numerical experiments in Section \ref{sec:castestudy} allow the application of hard constraints, we apply the model introduced in (\ref{eq:MIP_objective})-(\ref{eq:variable_domain_alpha}), as it proves to be more efficient in terms of runtime while guaranteeing request fulfillment.} 

\paragraph{Constructing Training Points.}
After deriving the full information solution spanning over the whole problem horizon $\mathcal{T}$, we are able to slice this solution into training instances. We visualize the rationale of our slicing approach in Figure~\ref{fig:instance_extraction}. At each time step within our time horizon $t \in \{0,...,T\}$, we extract the current system state $x_t$ at which a dispatch or redeployment decision is made. The decisions made in previous periods correspond to the optimal decisions extracted from the full information solution. The offline solution allows us to easily map each instance $x_t$ to its optimal decision $y'_t$ made at each time step $t$.
\begin{figure}[h!]
\centering
  \includegraphics[width=\linewidth, trim={0 .2cm 0 .0cm}, clip]{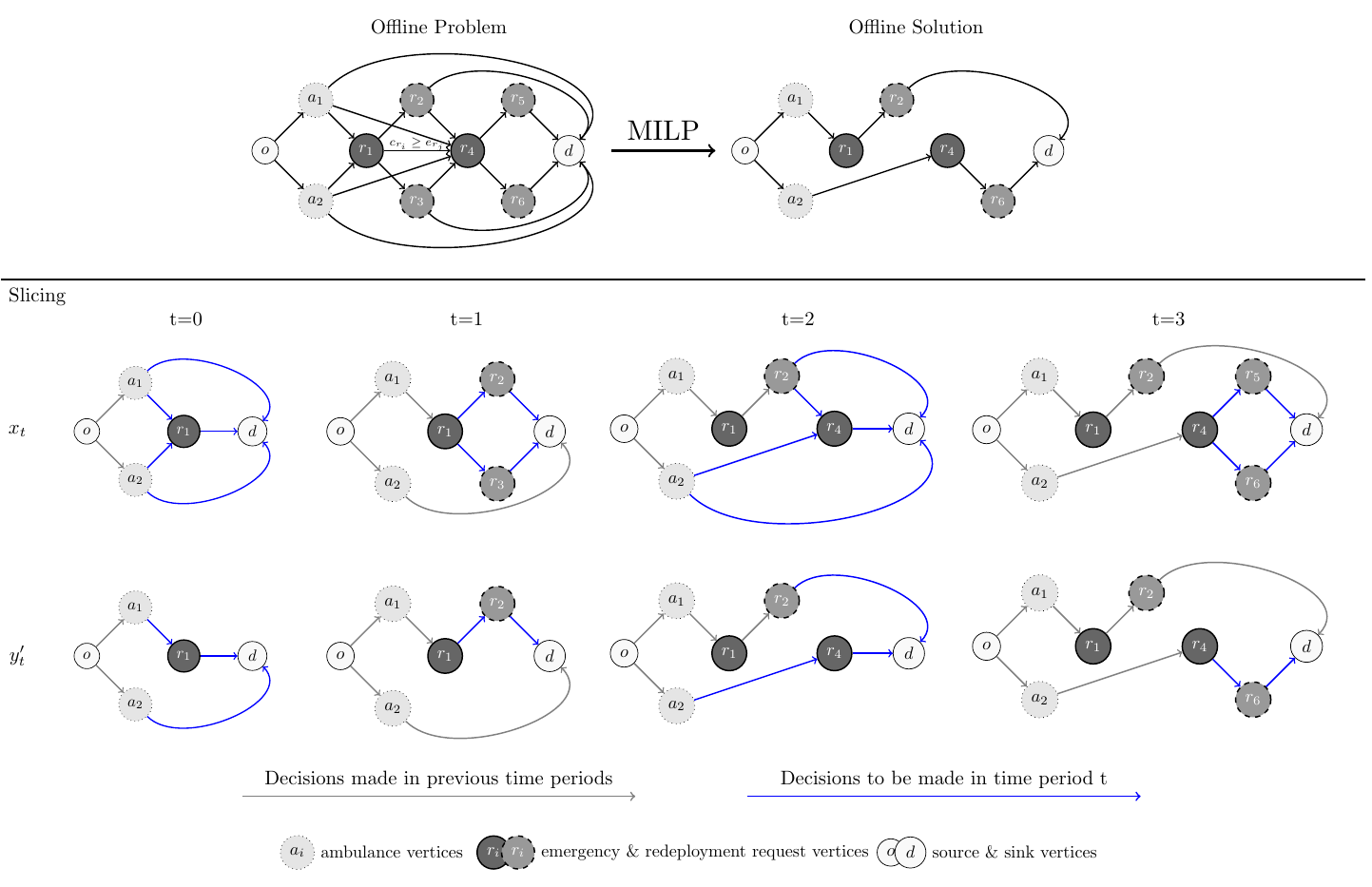}
  \caption{Offline Solution Slicing}
  \label{fig:instance_extraction}
\end{figure}

\subsection{Online Decision-Making: Designing a CO-augmented ML pipeline}\label{sec:methodology_COML_pipeline_design}
To take decisions ad-hoc when emergency requests arrive, we develop a CO-augmented ML pipeline as shown in Figure~\ref{fig:pipeline}. Developing such a pipeline requires defining a suitable statistical model for the ML-layer, as well as a suitable optimization problem for the CO-layer. In the following, we detail our design decisions for both layers.

\paragraph{ML-Layer.}
We apply a linear regression (LR) and a multilayer perceptron (MLP) model as predictors in the ML-layer to parameterize our CO model. The LR model represents a simple but easily interpretable statistical model:
\begin{equation}
    Y = \alpha + X\beta + \epsilon
\end{equation}
where $\alpha$, $\beta$, and $\epsilon$ are the y-intercept,  the regression coefficients, and the error term, respectively. The model benefits from its simplicity, making it easy to train, apply, and interpret. However, assuming linear relationships among the variables may limit its ability to capture more complex patterns within the data.

MLPs are fully connected feedforward neural networks consisting of an input layer, an output layer, and one or more hidden layers, which capture non-linearity. Each hidden layer $l$ computes the weighted sum of its inputs, adds a bias term $b^{l}$, and applies an activation function $\vartheta$. Formally:
\begin{align}
    h=\vartheta\Big(W_lx + b_l\Big),\label{eq:denselayer}
\end{align}
where $W_l$ denotes the layer's weights. In contrast to LR, MLPs can learn more complex, non-linear patterns, which may improve their performance. However, the strengths of the MLP come at the cost of higher computational requirements needed for tuning its hyperparameters and training. Also, decisions made by the MLP are typically less explainable than those made by simpler models such as LR. 

\paragraph{CO-Layer.}
In the online setting, our system evolves on an event basis: upon the arrival of emergency requests, we make dispatching decisions, and upon the completion of emergency requests, we make redeployment decisions. To take effective decisions in this setting, it is essential to choose a CO-layer that allows to capture the decision-making problem's structure but is at the same time efficiently solvable to obtain fast decisions.

To balance both objectives, we rely on a matching problem that naturally incorporates characteristics of the respective dispatching decisions and is at the same time efficiently solvable. Specifically, we model the optimization problem in our CO-layer as a weighted bipartite matching problem in which we match ambulances to dispatching or redeployment requests. We model the problem on a weighted bipartite graph $G = ({\mathcal{V}}^{\mathcal{M}}, {\mathcal{V}}^{\mathcal{R}}, {\mathcal{E}})$ where ${\mathcal{V}}^{\mathcal{M}}$ and ${\mathcal{V}}^{\mathcal{R}}$ are disjoint vertex sets representing a set of ambulances and a set of requests. Each edge $e \in {\mathcal{E}}$ links an ambulance to an emergency or redeployment request representing a feasible dispatching or redeployment decision.
We encode any feasible solution as a binary vector $y \in \{0, 1\}^{|{\mathcal{E}}|}$ where 1 indicates that the respective edge is chosen and 0 otherwise. To derive an online solution $\hat{y} \in \mathcal{Y}(x)$ for problem instance $x \in \mathcal{X}$, we leverage the edge parametrization $\theta$ derived from our ML-Layer. Given this edge parametrization, we aim at finding a matching $M$ that maximizes the cost $c$ of the matching, i.e., $c(M) = \theta^Ty$. Hence, we obtain our online solution by solving a weighted bipartite matching problem, formally
\begin{equation}
    \hat{y} = g(\theta) = \argmax_{y \in \mathcal{Y}(x)} \theta^Ty.
    \label{eq:learning_minimization_problem}
\end{equation}

\subsection{Structured-Learning Methodology}\label{sec:structured_learning_methodology}
Given a set of combinatorial training points $(x_i,y'_i)$, we aim to find a parameterization $w$ for our statistical model $\varphi_w$, such that for any given $x_i$, it outputs a parameterization $\theta$ that guides the bipartite matching in the CO-layer to find a solution $\hat{y}_i$ that aligns with the respective ground truth solution $y'_i$. Formally, we aim to solve the learning problem
\begin{equation}
    \min_{\theta} \frac{1}{n}\sum_{i=1}^{n} \mathcal{L}(\theta,x_i,y_i').\label{eq:learning_problem}
\end{equation}
Note that solving this learning problem allows to derive a parameterization $w$ for $\varphi_w$; as $\varphi_w: x_i \mapsto \theta$, we can derive a meaningful update for $w$ by building a gradient $\nabla_\theta\mathcal{L}$ over $\theta$.

Learning such a parameterization requires defining the right loss function and learning paradigm. In the following, we detail such a loss function and learning scheme, before enhancing it to improve our pipeline's generalization performance in a stochastic multi-stage decision-making setting.

\paragraph{Loss Function \& Learning.}
As we can describe a solution $\hat{y}_i$ as $\hat{y}_i = \arg\max_{y \in \mathcal{Y}(x_i)} \theta^T y$, a natural choice for a loss function is 
\begin{equation}
    \mathcal{L}(\theta, x, y') = \max_{y \in \mathcal{Y}(x)} \{ \theta^T y \} - \theta^{T}y',
\end{equation}
which allows to compute a surrogate measure for the suboptimality of $\hat{y}_i$ relative to $y_i'$: the closer the loss is to zero, the smaller is the difference between $\hat{y}_i$ and $y_i'$.
We compute this loss by subtracting the sum of weights induced by the optimal solution from the sum of weights induced by the learned solution. The maximization term on the first part of the right-hand side ensures that the computed gap always considers the difference between best achievable $\hat{y}_i$ under $\theta$ and the ground truth. Assuming the ground truth to be optimal (see Section~4.1), this gap always remains positive unless $\hat{y}_i$ matches $y_i'$ where the gap is zero. Hence, optimizing this loss allows us to adjust $\theta$, i.e., the parameterization $w$ of $\varphi_w: x_i \mapsto \theta$ to ensure that the CO-layer solution converges to the ground truth.

While this loss function remains intuitive from a CO perspective, it bears two fundamental problems from a learning perspective: first, the mapping \(\theta \rightarrow y'(\theta)\) is piecewise constant, with undefined or zero gradients across its domain. Second, the learning problem is degenerate: Setting $\theta=0$ leads to any feasible point in $\mathcal{Y}$ being an optimal solution, effectively reducing our learning problem to a random selection. To address these issues, we follow \cite{berthet2020learning} and perturb the loss:
\begin{equation}
    \mathcal{L}_\epsilon(\theta, x, y') = \mathbb{E}[\max_{y \in \mathcal{Y}(x)} \{ (\theta + \epsilon Z)^T y \}] - \theta^Ty'\label{eq:perturbed_loss}
\end{equation}
where the perturbations' amplitude is defined by $\epsilon > 0$ and $Z \sim \mathcal{N}(0, I)$ is a Gaussian vector. Considering the convex hull of the feasible solutions $\mathcal{Y}$, this perturbation induces a probability distribution over the vertices of the polytope. By introducing this perturbation, the loss function becomes smooth and convex.

Formally, one can give proof to these characteristics by leveraging Fenchel duality as in \citet{berthet2020learning}: we can show that our perturbed loss function corresponds to the left-hand side of the Fenchel-Young inequality 
\begin{equation} \Omega^*(\theta) + \Omega(y) - \theta^{\top} y \geq 0, 
\end{equation} 
where $\Omega$ is the Fenchel conjugate of $F(\theta) = \mathbb{E}(\max_y(\theta + \epsilon Z)^\top y)$. Since both $\Omega$ and its conjugate $\Omega^*$ are strictly convex, the learning problem remains strictly convex with a unique optimal solution that avoids degenerate cases. Specifically, $\mathcal{L}_\epsilon(\theta, x, y')$ attains zero only if $\theta$ matches the Fenchel conjugate of $y_i'$, which would be at an extreme value within the normal cone of $y_i'$ if $y_i'$ is a vertex.

Since the above summarized Fenchel-Young duality ensures that our perturbed loss remains convex and smooth \citep{berthet2020learning}, we obtain a meaningful gradient for our learning problem that reads 
\begin{equation}
    \nabla_\theta\mathcal{L}_\epsilon(\theta, x, y') = \mathbb{E}[ \argmax_{y \in \mathcal{Y}(x)} (\theta + \epsilon Z)^Ty ] - y'.
\end{equation}
Given this gradient, we minimize the learning problem in~(\ref{eq:learning_problem}) using loss~(\ref{eq:perturbed_loss}). Since computing the expectation in~Z is infeasible, we approximate it using a sample average applying stochastic gradient decent to parameterize $\varphi_w$.

\paragraph{Generalization.}
So far, we only derived an anticipative training set $\mathcal{D}$ which allows to apply the learning paradigm outlined above, but suffers from a fundamental disadvantage in a stochastic multi-stage optimization setting. In fact, training our pipeline on the anticipative training set $\mathcal{D}$ leads to a distribution mismatch problem. This kind of distribution mismatch problem is well known in standard supervised learning for sequential decision-making tasks, where a common approach is to train a policy using demonstrations from an expert (e.g., a human or an oracle). If the trained policy makes a mistake during inference, it may visit states that the expert never encountered during training. This results in compounding errors, where the policy drifts further from the expert's trajectory and encounters situations it cannot handle.

In our specific case, the distribution mismatch problem is as follows: all states that we consider during training are states that we observe when following an optimal policy $\pi'$ and the learning problem we solve reads
\begin{equation}
    \hat{w} = \min_{w} \mathbb{E}_{x \sim \pi'} \left[ \mathcal{L}\left(g(\varphi_w(x)), \pi'(x)\right) \right].
\end{equation}
However, to obtain a good generalization performance, we would like to solve the following learning problem
\begin{equation}\label{eq:learning_problem_w}
    \hat{w} = \min_{w} \mathbb{E}_{x \sim \pi_{w}} \left[ \mathcal{L}\left(g(\varphi_w(x)), \pi'(x)\right) \right],
\end{equation}
where we also visit states that are visited when applying non-optimal policies $\pi$. This allows to explore a larger share of the problem's transition kernel, which ultimately leads to a better performance when taking non-optimal decisions.

As such distribution mismatch problems have been frequently encountered in supervised learning settings, there exists a methodological paradigm to correct the mismatch, referred to as Dataset Aggregation (\texttt{DA\scriptsize{GGER}}). This technique was introduced by \cite{ross2011reduction} and relies on progressively collecting new training data from the policy’s own rollouts and labeling these with the expert’s decisions. However, applying \texttt{DA\scriptsize{GGER}} remains computationally heavy in our combinatorial setting as it requires to continuously regenerate the training points during learning \citep[cf.][]{greif2024combinatorial}. To mitigate this computational overhead, we propose an alternative dataset augmentation that can be performed apriori to the learning phase. In the following, we describe both the \texttt{DA\scriptsize{GGER}} paradigm and our alternative augmentation approach.

\textit{\texttt{DA\scriptsize{GGER}}:} Algorithm~\ref{alg:dagger} 
\begin{algorithm}
\caption{\texttt{DA\scriptsize{GGER}}}
\label{alg:dagger}
\begin{algorithmic}[1]
\State \textbf{Given:} dataset $\mathcal{D}=\emptyset$, reference policy $\pi'$, time horizon $\mathcal{T}$, initial state $x_0$, $w$ undefined, $\alpha = [\alpha^{[1]}, ..., \alpha^{[n]}]$
\For{$i = \{1,...,n\}$}
    \For{$t = \{0,...,T-1\}$}
        \State Transition to $x_{t+1}$ with (\ref{eq:transition}) where $y_{t} = \alpha^{[i]}\pi'(x_t)+(1-\alpha^{[i]})\pi_w^{[i]}(x_t)$
        \State $\mathcal{D} \leftarrow \mathcal{D} \cup \{(x_{t}, y_{t}')\}$
    \EndFor
    \State Update $w$ by solving (\ref{eq:learning_problem}) based on $\mathcal{D}$
\EndFor
\State Return best $w$ evaluated on the validation set
\end{algorithmic}
\end{algorithm} 
shows a pseudo-code of the  \texttt{DA\scriptsize{GGER}} routine. \texttt{DA\scriptsize{GGER}} iteratively builds a training set using a stochastic mixture policy, which combines the reference policy $\pi'$ and the learned policy $\pi_{w}$ by probabilistically selecting one of them for each transition (l.~4). In each iteration, the predictor is updated on the augmented training set (l.~7). Typically $\alpha$ decreases with each iteration, increasing the chance of transitioning with the learned policy. Constructing the training set iteratively during the learning process can be computationally expensive. To address this issue, we propose an alternative algorithm enabling the training set generation a priori in the following.

\textit{Enhanced Training Set $\mathcal{D}'$:} For building an improved training set $\mathcal{D}'$ prior to training, we make use of non-optimal policies $\tilde{\pi} \in \tilde{\Pi}$ to imitate the non-optimal decision-making induced by applying $\pi_{w}$. Algorithm \ref{alg:dataset} shows the pseudo-code. \begin{algorithm}
\caption{Enhanced dataset generation prior to training}
\label{alg:dataset}
\begin{algorithmic}[1]
\State \textbf{Given:} dataset $\mathcal{D}=\emptyset$, number of training instances to enhance the  dataset n, set of non-optimal policies $\tilde{\Pi}$, reference policy $\pi'$, time horizon $\mathcal{T}$, initial state $x_0$

\For{$t = \{0, ..., T-1\}$}
    \State Transition to $x_{t+1}$ with (\ref{eq:transition}) where $y_{t} = \pi'(x_t)$
    \State $\mathcal{D} \leftarrow \mathcal{D} \cup \{(x_{t}, y^{*}_{t})\}$
\EndFor

\State $\mathcal{D}' \leftarrow \mathcal{D}$

\For{$i = \{1,...,n\}$}
    \State Draw random timestamp $\tilde{t} \in \{t',...,T-1\}$ where $t' > 0$
    \For{$\tilde{\pi} \in \tilde{\Pi}$}
    
        \For{$t = \{1,...,\tilde{t}-1\}$}
            \State Transition to $x_{t+1}$ with (\ref{eq:transition}) where $y_{t} = \tilde{\pi}(x_t)$
        \EndFor
        
        \For{$t = \{\tilde{t}, ..., T-1\}$}
            \State Transition to $x_{t+1}$ with (\ref{eq:transition}) where $y_{t} = \pi'(x_t)$
        \EndFor

        \State $\mathcal{D}' \leftarrow \mathcal{D}' \cup \{(x_{\tilde{t}}, y_{\tilde{t}}')\}$
        
    \EndFor
\EndFor
\end{algorithmic}
\end{algorithm} As a basis, we generate the anticipative training set $\mathcal{D}$ by following the reference policy (ll.~2 - 5). Then, we repeatedly extend this training set for $n$ iterations as follows: First, we draw a random timestamp $\tilde{t} > 0$ within our time horizon (l.~8). Second, we transition by following a non-optimal policy up to timestamp $(\tilde{t} - 1)$ (ll.~10 - 12). Third, we solve the offline model for periods $[\tilde{t}, T]$, ensuring that we make an optimal decision at time $\tilde{t}$ (ll.~13 - 15). Fourth, we extract the instance and its optimal decision made at timestamp $\tilde{t}$, i.e., $(x_{\tilde{t}}, y_{\tilde{t}}')$, and add it to our enhanced training set $\mathcal{D}'$ (l.~16). We emphasize that one key difference between both approaches is that \texttt{DA\scriptsize{GGER}} (Algorithm \ref{alg:dagger}) augments the training set during the training process while we build the enhanced training set $\mathcal{D}'$ (Algorithm \ref{alg:dataset}) once prior to training. {\color{black}We compare the performance of both approaches in Section \ref{sec:casestudy_results}}.

\section{Case Study: Ambulance Demand Prediction for San Francisco}\label{sec:castestudy}
We conduct a numerical case study based on real data from San Francisco to evaluate the performance of the online policies learned by the presented ML pipeline and compare their performances to our benchmarks introduced in Section \ref{sec:benchmarks}. In Section \ref{sec:castestudy_setup}, we introduce the case study setup before presenting the results in Section \ref{sec:casestudy_results}.

\subsection{Current Industry Practice and Benchmarks}\label{sec:benchmarks}
In practice, dispatchers follow simple rules to dispatch and redeploy ambulances, which also serve as commonly applied benchmarks in the existing literature. We apply two online benchmarks detailed in following.
\begin{enumerate}
    \item Closest idle dispatching \& dynamic redeployment to the closest station (CICS)
    \item Closest idle dispatching \& static redeployment to a fixed station (CIFS)
\end{enumerate}

\paragraph{Dispatching.} For dispatching ambulances, we apply the closest-idle policy for both benchmarks. This policy always sends the nearest available ambulance to an incoming incident. It is commonly applied in practice \citep{aringhieri2017emergency}, and serves as a benchmark in literature \citep[see, e.g., ][]{schmid2012solving, jagtenberg2017optimal, liu2020ambulance, hua2022optimal}. 

\paragraph{Redeployment.} For redeployment, we apply a static policy and a dynamic policy. For the static policy, we determine a dedicated waiting position for each ambulance a priori, which the ambulance always returns to after serving a request. We refer to this station as an ambulance's \textit{home base}. This policy is often applied in practice, e.g., in Austria \citep{schmid2012solving}, and is also used as a common benchmark for redeployment models \citep[see, e.g., ][]{schmid2012solving, ji2019deep}. As the performance of this policy highly depends on the assignment of ambulances to home bases, we apply an extended version of the MCLP to allocate ambulances $m \in \mathcal{M}$ to stations $s \in \mathcal{S}$. We present the applied MCLP in Appendix \ref{sec:appendix_coverage_model}. In line with \cite{schmid2012solving} and \cite{ji2019deep}, we further benchmark against a dynamic policy, which always sends an ambulance to its closest waiting position after serving a request. In contrast to the dynamic redeployment policy, the selected waiting positions are not known a priori.

\subsection{Case Study Setup}\label{sec:castestudy_setup}
We base our experiments on San Francisco's 911 call data\footnote{https://data.sfgov.org/} considering calls of the category \textit{Medical Incidents} served by \textit{Advanced Life Support (ALS)} units. In our study, all ambulance stations 
serve as possible waiting positions for idle ambulances. We further assume that patients are transported to the nearest emergency room 
if a transport is required. In line with \cite{rautenstrauss2023ambulance}, we calculate driving times, assuming a driving velocity of 30 km/h, accounting for traffic and the Haversine distance. We base the locations of ambulance stations and emergency rooms on real data and visualize them in Figure~\ref{fig:resources}.

\begin{figure}[h!]
\centering
  \includegraphics[width=.5\linewidth]{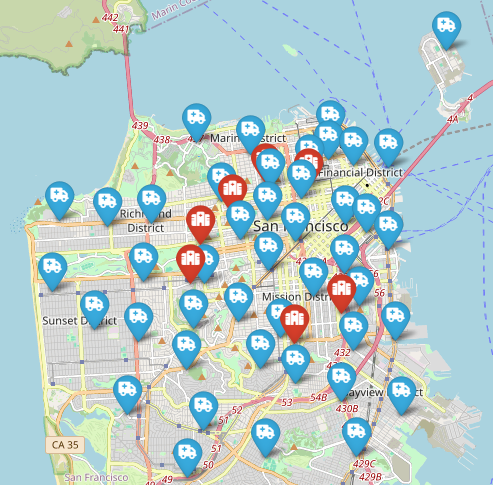}
  \caption{Locations of Emergency Rooms (red) and Ambulance Stations (blue)}
  \label{fig:resources}
\end{figure}

We evaluate the performance of the learned policies for various scenarios. First, we vary the request density by applying the policies during two different time periods of the day: 12 a.m. to 6 a.m., representing a low-demand scenario with an average number of 36.7 emergency incidents, and 12 p.m. to 6 p.m., representing a high-demand scenario, with an average number of 71.6 emergency incidents, correspondingly. Figure~\ref{fig:temporal-pattern} visualizes the temporal characteristics of the emergency call patterns.
\begin{figure}[h!]
\centering
  \includegraphics[width=.7\linewidth, trim={0 .2cm 0 .0cm}, clip]{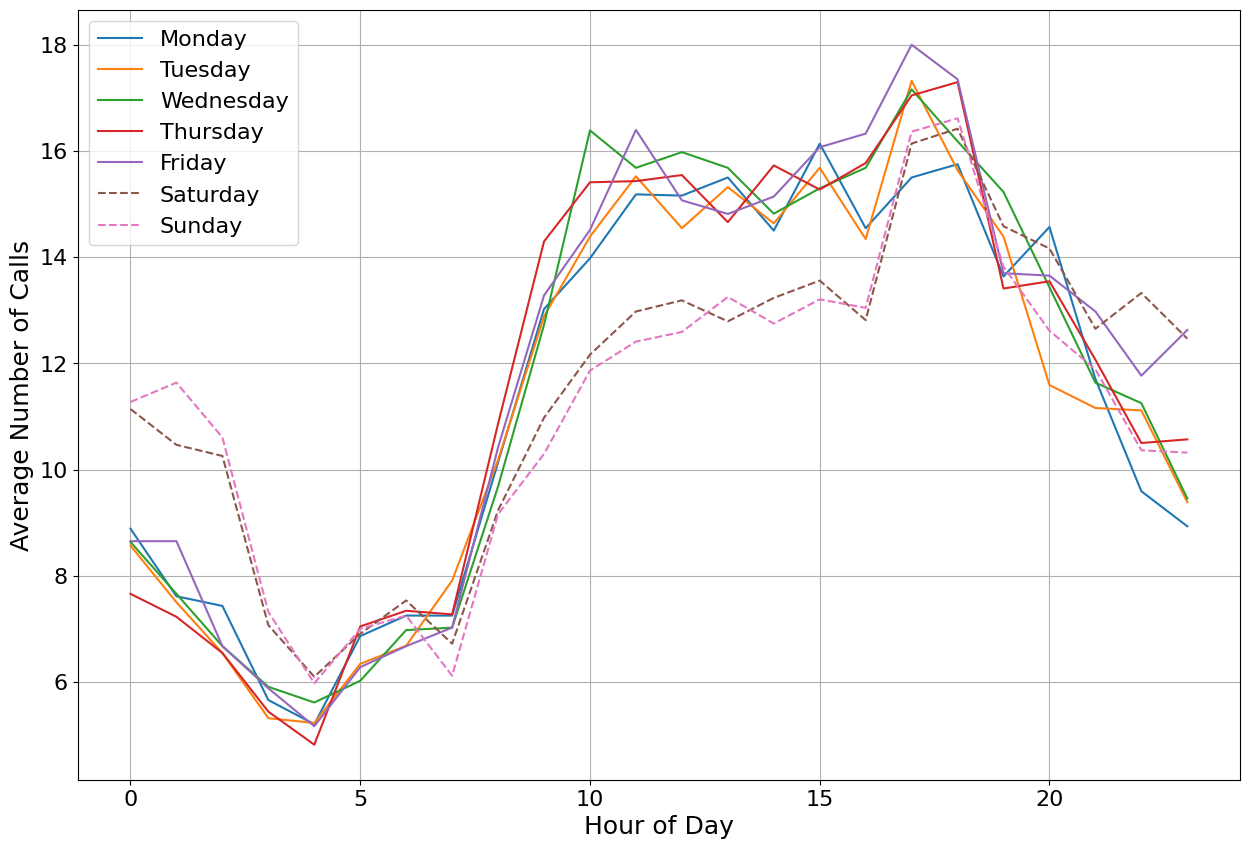}
  \caption{Temporal Emergency Call Patterns}
  \label{fig:temporal-pattern}
\end{figure}
For evaluation, we exclude the first hour of the investigated period, allowing the system to initialize and avoid starting in an idle state. Additionally, we exclude the last hour to mitigate a potential end-of-horizon effect, where ambulance redeployment decisions lose relevance. Second, the emergency call patterns observed on weekdays differ from those observed on weekends (see Figure~\ref{fig:temporal-pattern}). For this reason, we investigate the performance of the learned policies when training dedicated models only on weekdays and weekends. Third, we examine two resource scenarios in which we vary the number of ambulances: one with 25 ambulances and the other with 50 ambulances. The choice of 50 ambulances reflects the real-world situation observed in our data. The 25-ambulance scenario allows us to explore a case with limited resources, providing insights into the system's performance under more constrained conditions.

\paragraph{Training.}
We train both predictors on two different training sets: i) The \textit{anticipitative training set} $\mathcal{D}$ containing 250 randomly sampled instances from 7 days (4th October - 10th October) induced by transitioning with the optimal policy and ii) the \textit{enhanced training set} $\mathcal{D}'$ containing 170 randomly sampled instances from 7 days (4th October - 10th October) induced by transitioning with the optimal policy and 80 randomly sampled instances induced by transitioning with a non-optimal policy. We refer to Section~\ref{sec:structured_learning_methodology} for details on how we construct both training sets. We use 7 days (11th October - 17th October 2023) of data for validation and 14 days (18th October - 31st October 2023) for testing. We train the models for 250 epochs on an Intel(R) Xeon(R) processor E5-2697 v3 with 56 GB RAM. We apply the Adam optimizer and a learning rate of 0.001 for both the LR model and the MLP, optimizing their hyperparameters via grid search. Table \ref{table:features} presents the used features. We exclude features with a correlation of $>80\%$. No further feature selection was conducted, as the perturbation during training functions as a regularization mechanism. By inducing noise through perturbation, the model inherently reduces the relevance of less important features, preventing the model from overfitting. For evaluation, we utilize the first hour as a warmup period during which we apply the optimal dispatch and redeployment strategy. This provides an equal initial system state for all policies, enabling a fair comparison.
\begin{table}[h!]
\centering
\begin{adjustbox}{width=\textwidth}
\begin{tabular}{l | l }

Description&Features\\ \hline\hline

\multirow{9}{*}{Features for redeploying ambulances}
&Nr. of idle vehicles at redeployment location\\
&Nr. of idle vehicles within 1*/2*/3 km of redeployment location\\
&Nr. of vehicles traveling to redeployment location\\
&Nr. of vehicles traveling to redeployment location within 1*/2/3 km\\
&Nr. of busy vehicles with known drop-off location within 1/2/3* km of redeployment location\\
&Nr. of busy vehicles with unknown drop-off location operating within 1/2/3 km of redeployment location\\
&Nr. of requests within 1 km of redeployment location in the past 1/8/24*/48* hours\\
&Nr. of requests within 2/3 km of redeployment location in the past 1/8/24/48 hours*\\
&No idle vehicle at redeployment location (binary)\\ 
&Reciprocal distance between ambulance and redeployment location\\ \hline

\multirow{9}{*}{\shortstack{Features for dispatching ambulances \\ at/traveling to a waiting position}}
&Nr. of idle vehicles at waiting position\\ 
&Nr. of idle vehicles at waiting position within 1/2/3 km*\\ 
&Nr. of vehicles traveling to waiting position\\ 
&Nr. of vehicles traveling to waiting position within 1*/2/3 km\\
&Nr. of busy vehicles with known drop-off location within 1/2/3 km of waiting location\\
&Nr. of busy vehicles with unknown drop-off location operating within 1/2/3* km of waiting position\\
&Vehicle is traveling to/idling at waiting position (binary)\\
&Nr. of requests within 1 km of waiting position in the past 1/8/24*/48* hours\\
&Nr. of requests within 2/3 km of waiting position in the past 1/8/24/48 hours*\\
&Reciprocal distance between waiting position and unserved request\\
\hline

\multirow{2}{*}{Features for direct dispatches}
&Drop-off location is known (binary) x reciprocal distance\\
&Drop-off location is unknown (binary)\\

\end{tabular}
\end{adjustbox}
\caption{Features (*Excluded based on Correlation Analysis)}
\label{table:features}
\end{table}

\section{Results}\label{sec:casestudy_results}

In this section, we present the results of our numerical case study, comparing the mean response times achieved by the learned policies against those of the benchmark policies. First, we analyze the performance of the pipeline trained on daily data, which includes both weekdays and weekends. Second, we examine the effectiveness of our pipelines specifically trained for weekdays and weekends. Third, we provide a structural analysis of the different dispatching decisions made by both the benchmark and the learned policies. Finally, we compare the generalization performance achieved by training the pipeline on the {enhanced data set} against the \texttt{DA\scriptsize{GGER}} algorithm.

\subsection{Baseline Results}
\begin{figure}[!h]
\centering
\begin{subfigure}{.5\textwidth}
  \centering
  \includegraphics[width=.9\linewidth]{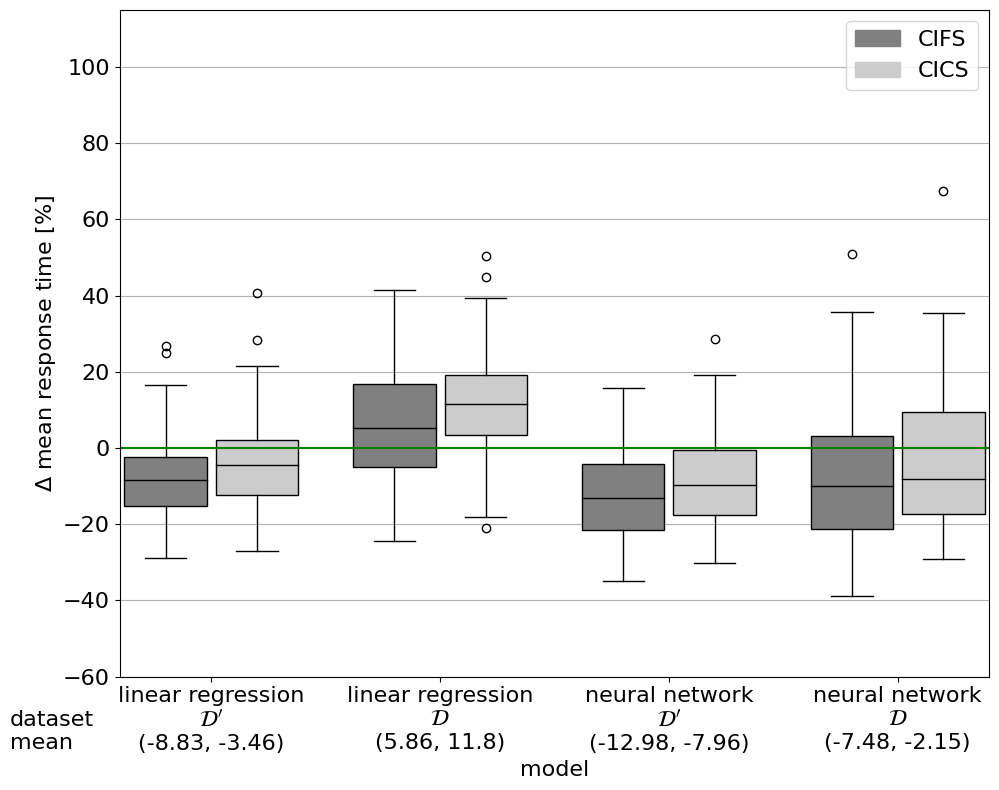}
  \caption{25 Ambulances, Low Demand}
  \label{fig:boxplot_response_time_daily_25v_low}
\end{subfigure}%
\begin{subfigure}{.5\textwidth}
  \centering
  \vspace{-0.2cm}
  \includegraphics[width=.9\linewidth]{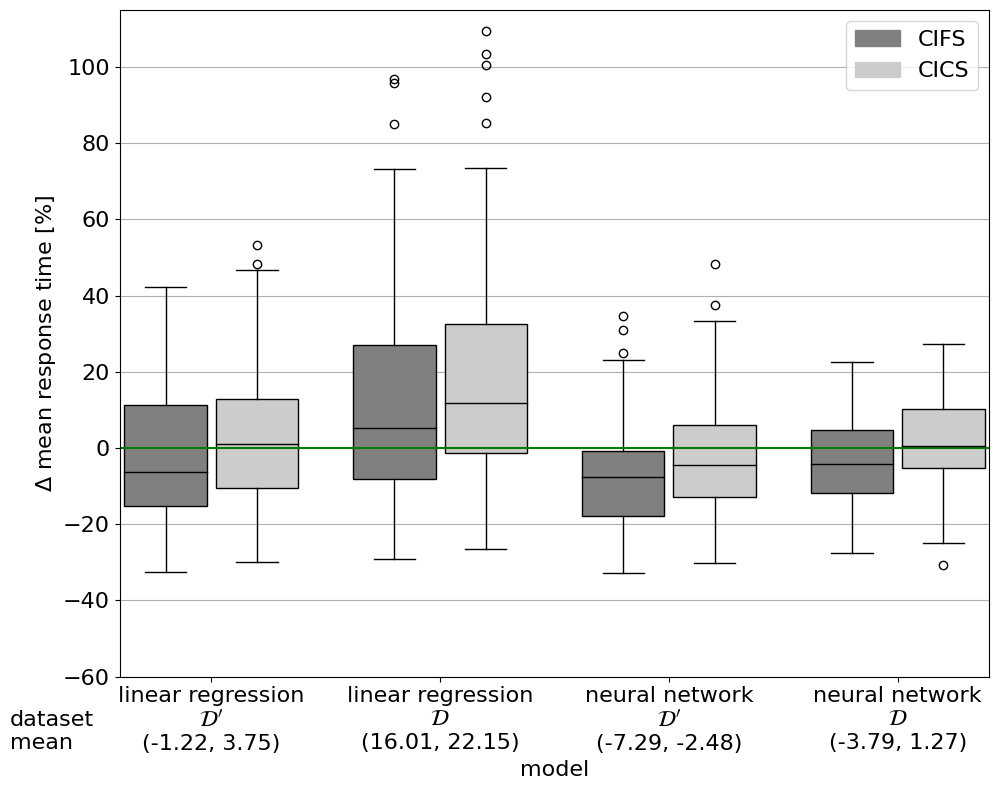}
  \caption{25 Ambulances, High Demand}
  \label{fig:boxplot_response_time_daily_25v_high}
\end{subfigure}

\begin{subfigure}{.5\textwidth}
  \centering
  \includegraphics[width=.9\linewidth]{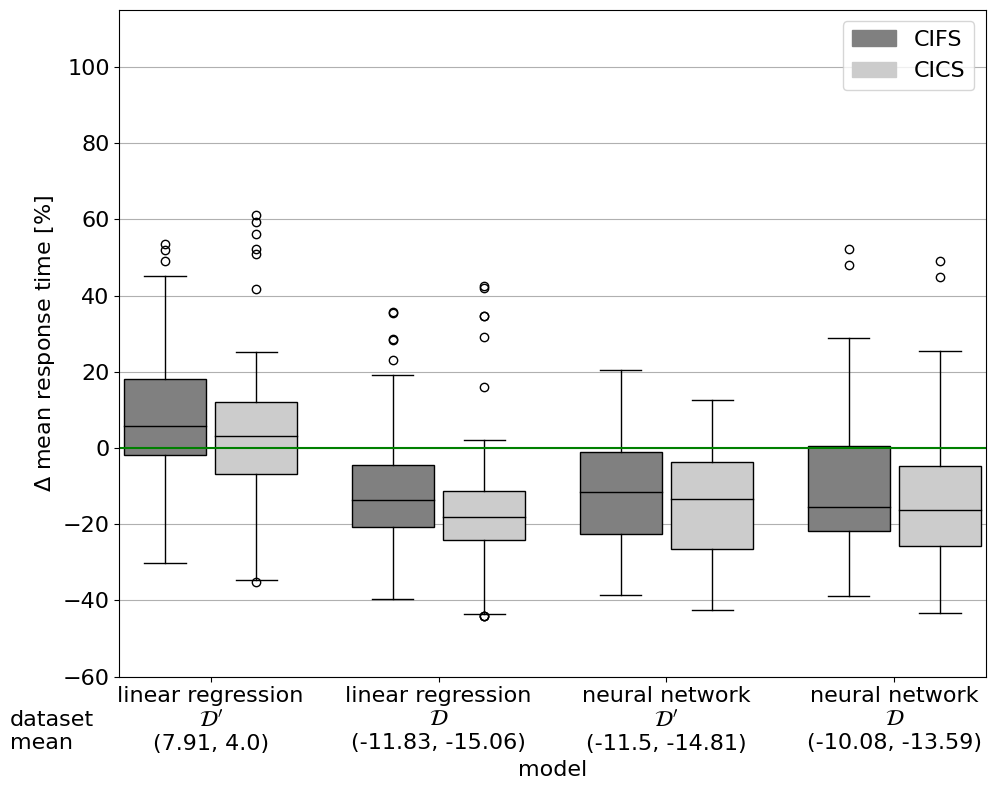}
  \caption{50 Ambulances, Low Demand}
  \label{fig:boxplot_response_time_daily_50v_low}
\end{subfigure}%
\begin{subfigure}{.5\textwidth}
  \centering
  \vspace{-0.2cm}
  \includegraphics[width=.9\linewidth]{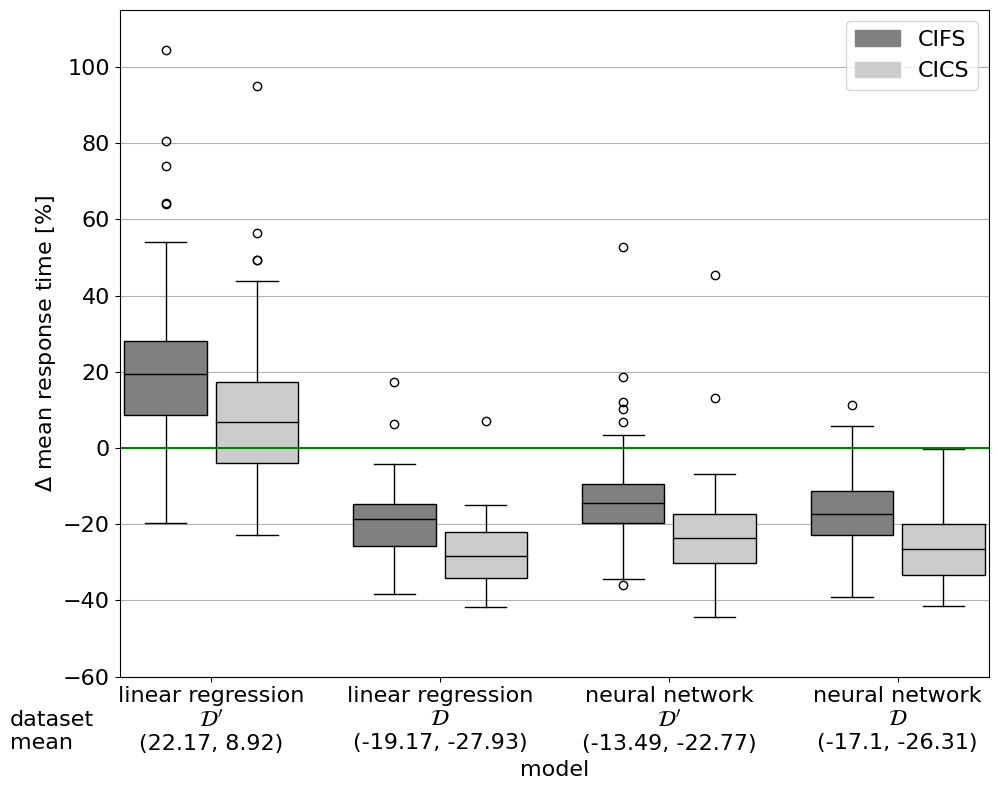}
  \caption{50 Ambulances, High Demand}
  \label{fig:boxplot_response_time_daily_50v_high}
\end{subfigure}
\caption{Model Comparison based on their Mean Response Time}
\label{fig:boxplot_response_time_daily}
\end{figure}
The plots in Figure~\ref{fig:boxplot_response_time_daily} show the percentage deviation in the mean response time when comparing the learned policies against our online benchmarks.   
For the scenarios operating 25 ambulances, we observe that the neural network consistently outperforms the LR model for both training sets. Additionally, enhancing the training set further improves the policies' performance. For the low-demand scenario (Figure \ref{fig:boxplot_response_time_daily_25v_low}), we observe a mean response time decrease of 12.98\% and 7.96\% over the online benchmarks applying the static and dynamic redeployment strategies, respectively. For the high-demand scenario (Figure \ref{fig:boxplot_response_time_daily_25v_high}), the mean response times can be reduced by 7.29\% and 2.48\%, respectively. The superiority of the neural networks trained on the enhanced data set can be explained by the increased complexity of decision-making when fewer ambulances are available. In low-resource scenarios, a single suboptimal dispatching or redeployment decision results in a higher risk of leaving regions uncovered. This leads to a higher risk of having longer response times compared to high-resource settings. The neural network, being better at capturing complex patterns, is superior in handling this complexity. Moreover, including states in the training set that result from suboptimal decision-making enhances the policies' robustness in low-resource situations. By exposing the model to scenarios induced by suboptimal policies during training, the learned policies are able to improve decision-making after a suboptimal decision has been made.

For the scenarios operating 50 ambulances (Figures~\ref{fig:boxplot_response_time_daily_50v_low} and  \ref{fig:boxplot_response_time_daily_50v_high}), most learned policies outperform both online benchmarks. The static redeployment policy is outperformed by up to 19\%, and the dynamic redeployment policy is outperformed by up to 28\%. Solely training an LR model on the enhanced data set reduces the performance by 8\% and 4\% compared to the static and dynamic deployment policy. This indicates that this model is too simple to learn a robust online policy on a dataset containing many scenarios induced by suboptimal decision-making, particularly when applied to a high-dimensional state space resulting from operating 50 vehicles. In general, contrary to the low-resource scenarios, the results show that training the models on the anticipative training set is superior to training the models on the enhanced data set. This can be explained by the fact that in the high-resource scenario, making a suboptimal dispatching decision is not as severe as in the low-resource scenario, as regions are mostly covered by more vehicles. For all scenarios presented in Figure~\ref{fig:boxplot_response_time_daily}, we show the average utilization of ambulances, i.e., the average percentage share of time an ambulance was busy, in Table~\ref{table:utilization}. The significantly lower utilization in high-resource scenarios shows that more ambulances are idling that are available to be dispatched.
\begin{table}[h!]
\centering
\begin{adjustbox}{width=.8\textwidth}
\begin{tabular}{lc|c|c|c|c}
 && \multicolumn{2}{c|}{{Low Demand}} & \multicolumn{2}{c}{High Demand} \\
 Model&Dataset& 25 Ambulances & 50 Ambulances & 25 Ambulances & 50 Ambulances \\\hline
 \multirow{2}{*}{{Linear Regression}}
 &$D$ & 26.82 & 13.36 & 58.67 &   28.05 \\
 &$D'$ & 25.72 & 12.52 & 57.84  & 27.89 \\\hline
 \multirow{2}{*}{{Neural Network}}
&$\mathcal{D}$ & 26.56 & 13.91 & 58.82  & 28.09  \\
&$\mathcal{D}'$ & 25.39 & 12.96 & 57.43  & 27.61  \\\hline
\end{tabular}
\end{adjustbox}
\caption{Average Percentage Share of Time Ambulances are Busy [\%]}
\label{table:utilization}
\end{table}

{\result{The learned policies outperform the online benchmarks by up to 19\% when redeploying ambulances to fixed stations and by up to 28\% when redeploying ambulances to the closest station in the high-resource scenario.}} 
{\result{The neural network trained on the enhanced training set $\mathcal{D}'$ consistently outperforms both online benchmarks across all scenarios when operating 25 ambulances. Enhancing the training set proves especially beneficial in resource-scarce scenarios, where decision-making becomes more complex.}}

\subsection{Training Dedicated Policies.} Given observed differences in the temporal emergency call patterns for weekdays and weekends as visualized in Figure~\ref{fig:temporal-pattern}, we investigate whether learning dedicated policies specifically for weekdays and weekends can further reduce the mean response time. In Table~\ref{table:vehicle_request_ratio}, we present the ambulance-request ratio for all settings, which shows the average number of requests an ambulance serves in each setting.
\begin{table}[h!]
\centering
\begin{adjustbox}{width=.8\textwidth}
\begin{tabular}{ll|c|c|c|c}
 && \multicolumn{2}{c|}{{Low Demand}} & \multicolumn{2}{c}{High Demand} \\
 Model&Dataset& 25 Ambulances & 50 Ambulances & 25 Ambulances & 50 Ambulances \\\hline
\multirow{2}{*}{{Dataset}} 
 & {Weekdays} & 1.30 & {0.65} & {3.00}  & 1.50 \\
 & {Weekends} & 1.86 & 0.93  & 2.73 & 1.36 \\\hline
\end{tabular}
\end{adjustbox}
\caption{Mean Number of Requests served per Ambulance}
\label{table:vehicle_request_ratio}
\end{table}

Our analysis reveals that the emergency call volume during weekend nights exceeds that of weekday nights by 43.01\%. Conversely, weekday afternoons see a 9.90\% higher volume of requests compared to weekends.

\paragraph{Dedicated Weekend Policies.} We visualize the results for dedicated weekend policies in Figure~\ref{fig:boxplot_response_time_weekends}. When evaluating the policies' performances on weekends, our results show that the models trained on daily data, which includes both weekdays and weekends, consistently outperform the models trained exclusively on weekend data. We observe performance improvements of up to 9.45 percentage points. The results indicate that the models trained solely on weekend data may overfit and can benefit from including daily data in the training process.  

\begin{figure}[!h]
\centering
\begin{subfigure}{.45\textwidth}
  \centering
  \includegraphics[width=.9\linewidth]{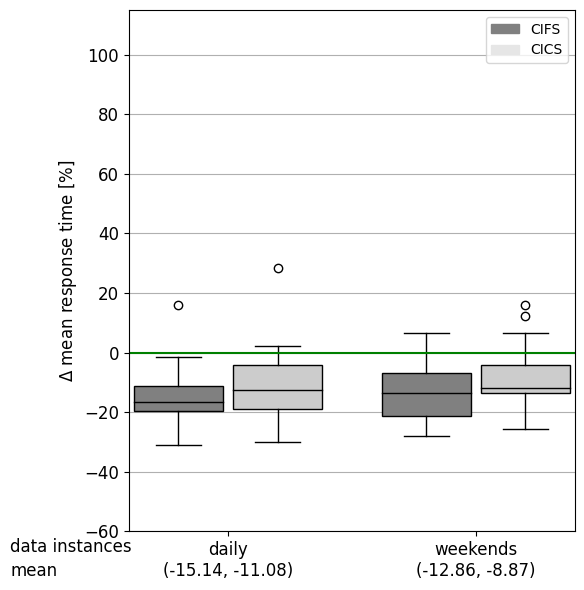}
  \caption{25 Ambulances, Low Demand}
  \label{fig:boxplot_response_time_weekdays_25v_low}
\end{subfigure}%
\begin{subfigure}{.45\textwidth}
  \centering
  \vspace{-0.2cm}
  \includegraphics[width=.9\linewidth]{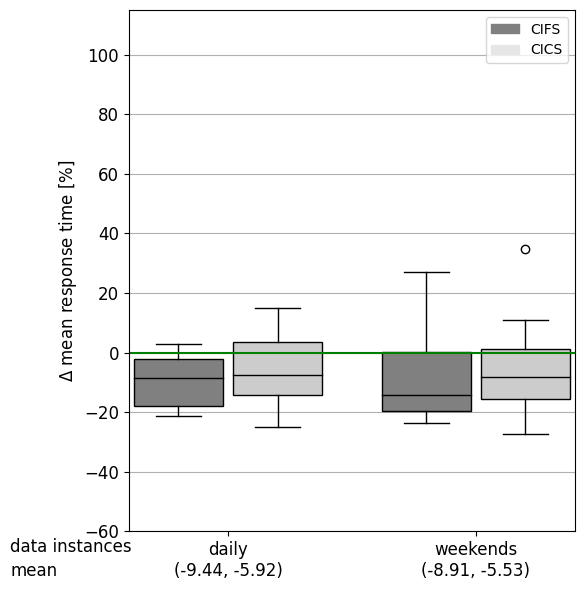}
  \caption{25 Ambulances, High Demand}
  \label{fig:boxplot_response_time_weekdays_25v_high}
\end{subfigure}

\begin{subfigure}{.45\textwidth}
  \centering
  \includegraphics[width=.9\linewidth]{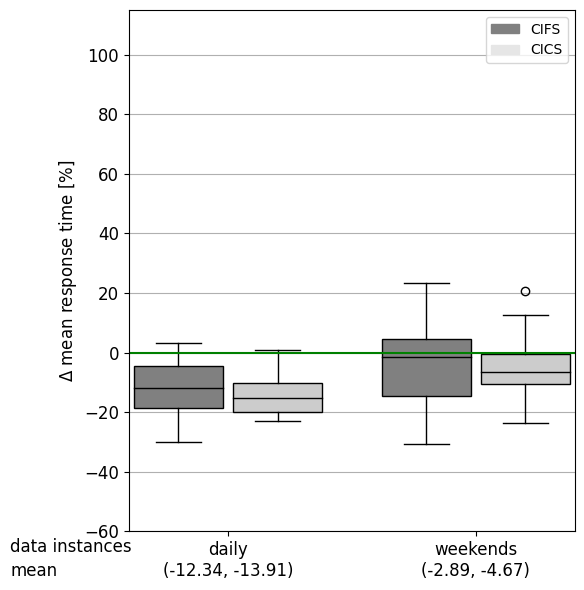}
  \caption{50 Ambulances, Low Demand}
  \label{fig:boxplot_response_time_weekdays_50v_low}
\end{subfigure}%
\begin{subfigure}{.45\textwidth}
  \centering
  \vspace{-0.2cm}
  \includegraphics[width=.9\linewidth]{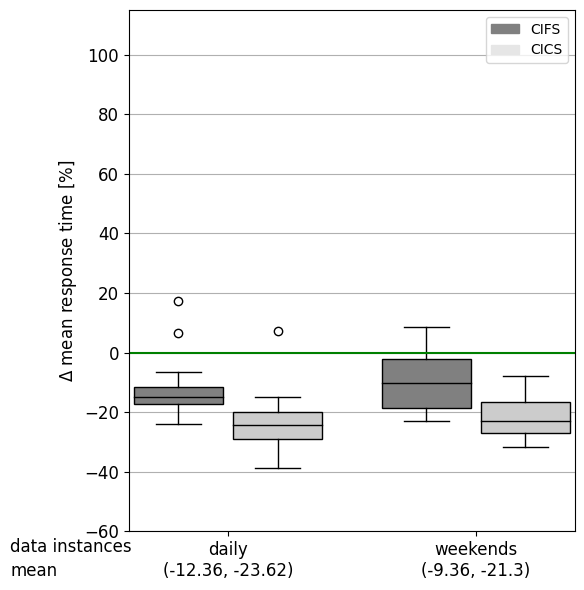}
  \caption{50 Ambulances, High Demand}
  \label{fig:boxplot_response_time_weekdays_50v_high}
\end{subfigure}
\caption{Percentage Difference in Mean Response Time - Dedicated Weekend Model}
\label{fig:boxplot_response_time_weekends}
\end{figure}

{\result{The model trained on daily data, encompassing both weekdays and weekends, consistently surpasses the model trained solely on weekend data by up to 9.45 percentage points when evaluating weekend performance.}}

\paragraph{Dedicated Weekday Policies.} We visualize the results for dedicated weekday policies in Figure~\ref{fig:boxplot_response_time_weekdays}. 

\begin{figure}[!h]
\centering
\begin{subfigure}{.45\textwidth}
  \centering
  \includegraphics[width=.9\linewidth]{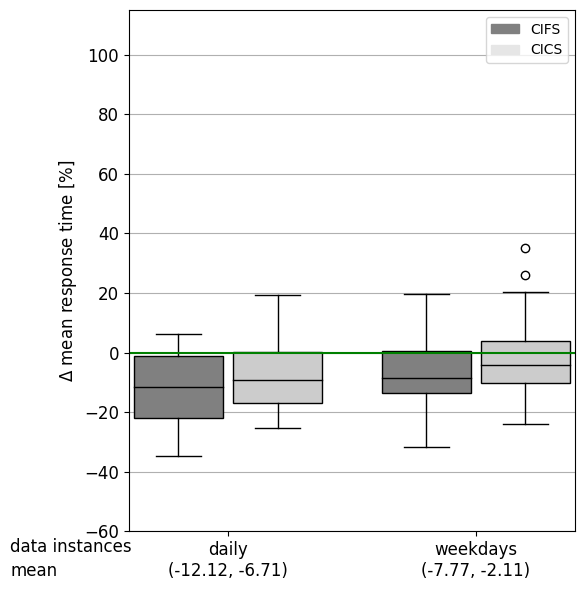}
  \caption{25 Ambulances, Low Demand}
  \label{fig:boxplot_response_time_weekdays_25v_low}
\end{subfigure}%
\begin{subfigure}{.45\textwidth}
  \centering
  \vspace{-0.2cm}
  \includegraphics[width=.9\linewidth]{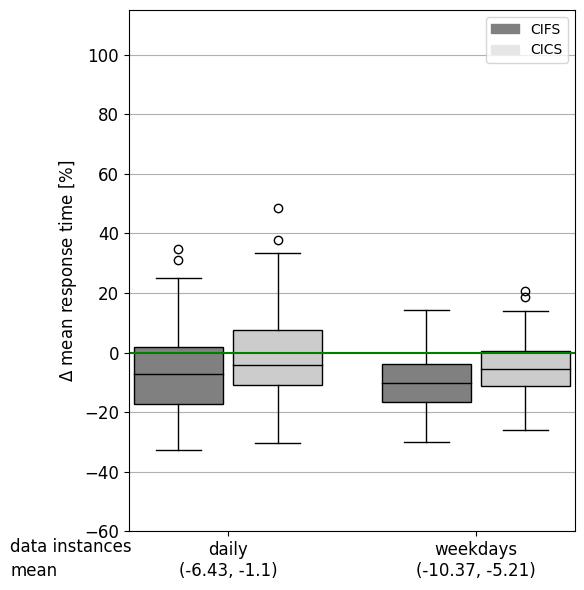}
  \caption{25 Ambulances, High Demand}
  \label{fig:boxplot_response_time_weekdays_25v_high}
\end{subfigure}

\begin{subfigure}{.45\textwidth}
  \centering
  \includegraphics[width=.9\linewidth]{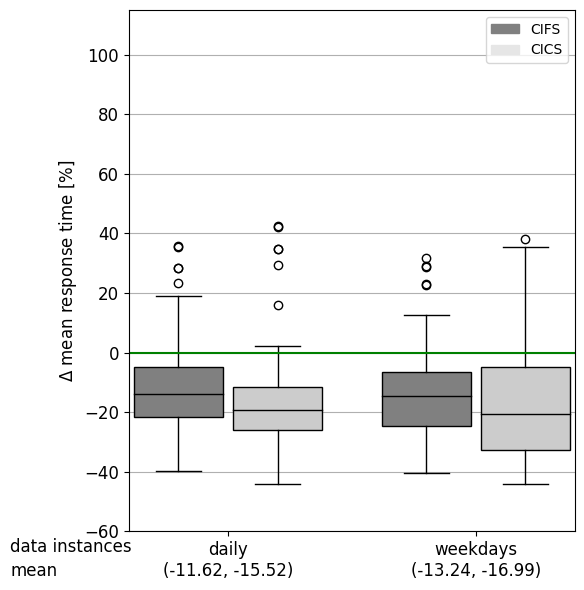}
  \caption{50 Ambulances, Low Demand}
  \label{fig:boxplot_response_time_weekdays_50v_low}
\end{subfigure}%
\begin{subfigure}{.45\textwidth}
  \centering
  \vspace{-0.2cm}
  \includegraphics[width=.9\linewidth]{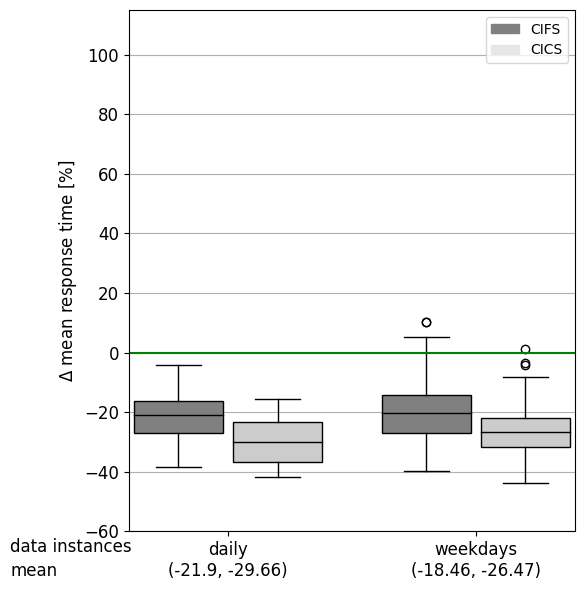}
  \caption{50 Ambulances, High Demand}
  \label{fig:boxplot_response_time_weekdays_50v_high}
\end{subfigure}
\caption{Percentage Difference in Mean Response Time - Dedicated Weekday Model}
\label{fig:boxplot_response_time_weekdays}
\end{figure}

When evaluating the policies' weekday performances, we see that in two scenarios, training the models on daily data is superior. Only for the scenarios with the highest and lowest ambulance-request ratios, we see that training a weekday-specific model can slightly outperform the models trained on daily data. However, these improvements are relatively small ranging between 1.47 and 3.94 percentage points. In general, we see that training the models on daily data mostly outperforms dedicated models. This indicates that daily models, in general, achieve a slightly more robust performance.

To gain a deeper understanding of these results, we conduct Kolmogorov-Smirnov tests evaluating the similarity between spatial emergency call patterns across weekdays and weekends. We provide details for these tests in Appendix \ref{sec:KS_test}. The results show that the spatial emergency call distributions observed on weekends differ from weekday distributions. This indicates that including data from all days reduces the chance of overfitting as the models learn more generalized patterns. We observe that the variation across weekdays is higher than the variation across weekends. This indicates that including data from different days is particularly important for weekend applications, where the data shows less variation compared to weekdays. 

{\result{Training the models on daily data outperforms dedicated weekday- or weekend-specific models. Including data from all days enables the models to better generalize, reducing the chance of overfitting.}}

\subsection{Policy Insights.} In Figure~\ref{fig:dispatching_trips}, we compare the dispatching decisions made by a learned policy (\ref{fig:trips_learned_policy}), the optimal policy (\ref{fig:trips_optimal_policy}), and the two online benchmark policies: closest idle dispatching and redeployment to either a fixed station (\ref{fig:trips_base_station}) or the closest station (\ref{fig:trips_closest_station}) for one night (12:00 a.m. - 06:00 a.m.) when operating 50 vehicles.
\begin{figure}[!h]
\centering
\begin{subfigure}{.23\textwidth}
  \centering
  \includegraphics[width=\linewidth]{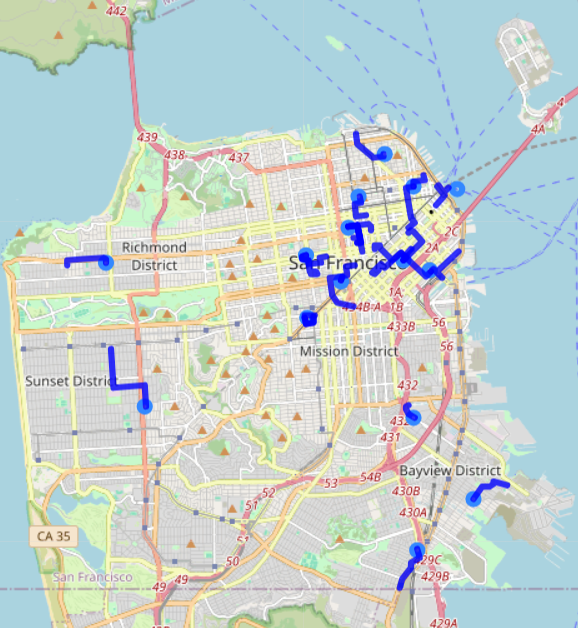}
  \caption{Learned policy}
  \label{fig:trips_learned_policy}
\end{subfigure}
\begin{subfigure}{.23\textwidth}
  \centering
  \includegraphics[width=\linewidth]{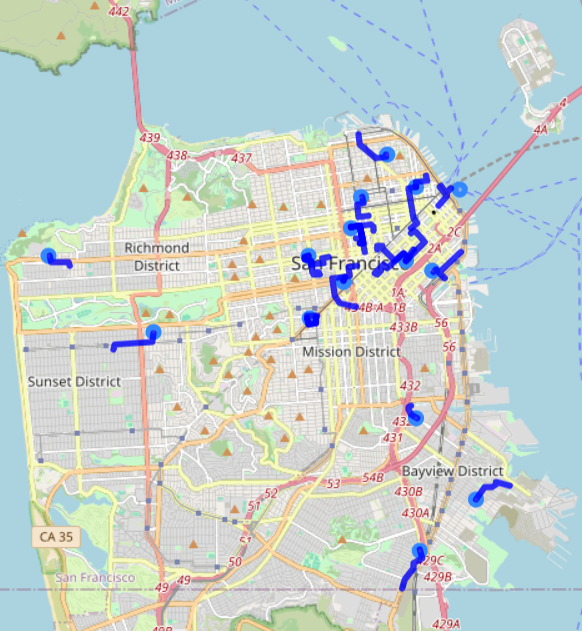}
  \caption{Optimal policy}
  \label{fig:trips_optimal_policy}
\end{subfigure}
\begin{subfigure}{.23\textwidth}
  \centering
  \includegraphics[width=\linewidth]{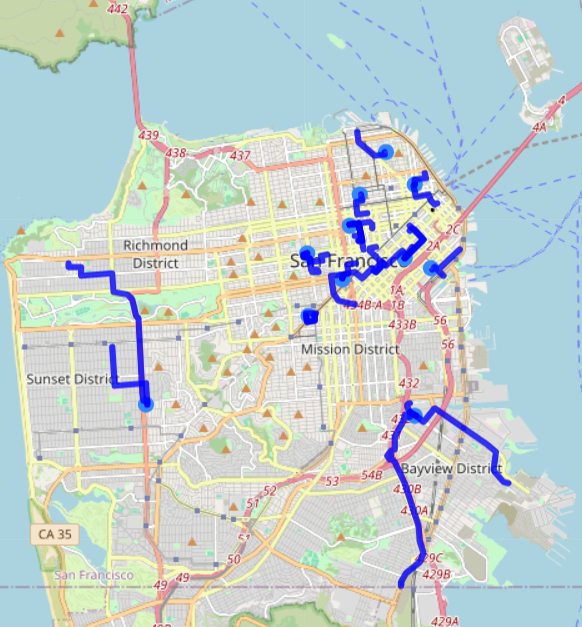}
  \caption{Fixed station}
  \label{fig:trips_base_station}
\end{subfigure}
\begin{subfigure}{.23\textwidth}
  \centering
  \includegraphics[width=\linewidth]{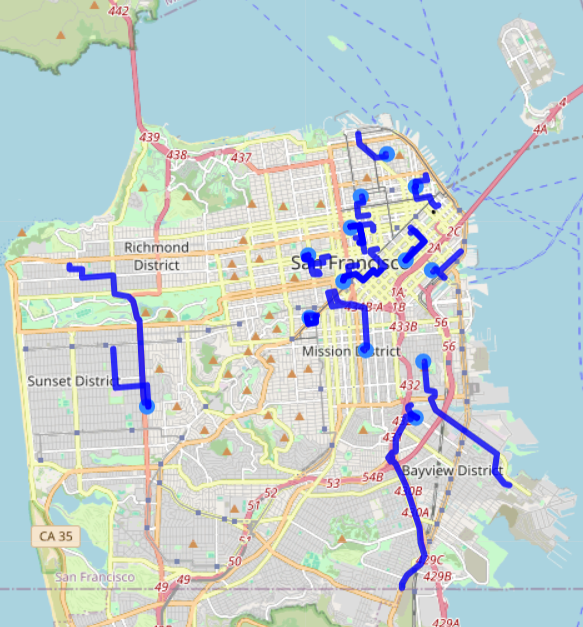}
  \caption{Closest station}
  \label{fig:trips_closest_station}
\end{subfigure}
\caption{Comparison of Dispatching Decisions}
\label{fig:dispatching_trips}
\end{figure}
The figure illustrates the routes taken by vehicles from their locations at the time of dispatch to the assigned incident locations. Notably, the learned policy leverages anticipatory redeployment decisions, enabling shorter dispatching routes and, consequently, reduced response times. This improvement is particularly evident in low-demand areas, such as the west and south of the region under study. The benchmark strategies, by contrast, either pull ambulances towards high-demand zones (in the case of the closest redeployment strategy) or assign them to high-demand areas in advance (in the static redeployment strategy), resulting in less efficient dispatches in low-demand regions.

{\result{The learned policy outperforms online benchmarks by making anticipatory redeployment decisions, reducing travel distances and response times, especially in low-demand areas.}}

\subsection{Training on Enhanced Data Set $\mathcal{D}'$ vs. \texttt{DA\scriptsize{GGER}}.} We propose an alternative approach to the \texttt{DA\scriptsize{GGER}} algorithm \citep{ross2011reduction} in Section \ref{sec:structured_learning_methodology}. 
The key differences between the two approaches are as follows: First,  \texttt{DA\scriptsize{GGER}} requires iterative updates during training, generating new instances and retraining the predictor in each iteration. In contrast, our method conducts a one-time data set augmentation before training. Second, while \texttt{DA\scriptsize{GGER}} augments the training set with instances derived from following the learned policy, our approach augments the training set with instances derived from following various dispatching and redeployment policies. This enables us to generate additional instances prior to training. Section \ref{sec:structured_learning_methodology} details both methods. To compare their performance, we evaluate both approaches using a neural network predictor, chosen for its superior performance over both online benchmarks in previous experiments (see Figures \ref{fig:boxplot_response_time_daily}–\ref{fig:boxplot_response_time_weekdays}). Due to the high computational cost of \texttt{DA\scriptsize{GGER}}, we restrict the comparison to low-demand scenarios. Figure~\ref{fig:dagger_comparison} presents the mean response time improvements over both online benchmarks and the total runtimes of the compared approaches. 

\begin{figure}[!h]
\centering
\begin{subfigure}{.5\textwidth}
  \centering
  \includegraphics[width=.95\linewidth]{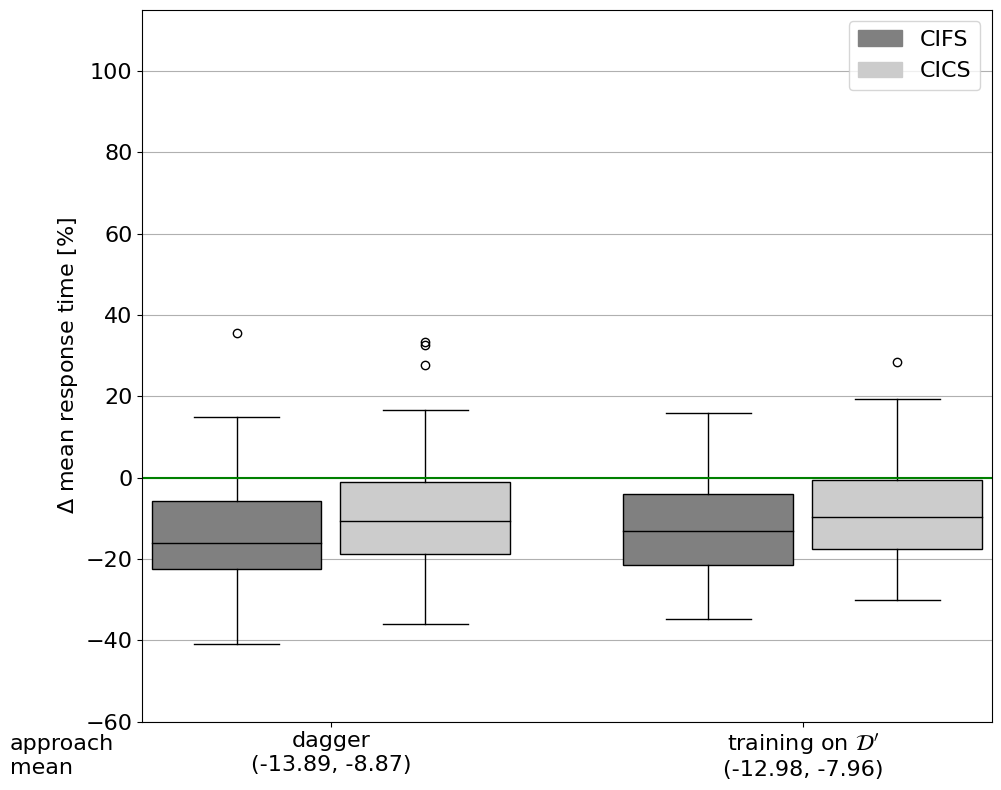}
  \caption{Mean Response Time Improvements: 25 Vehicles}
  \label{fig:boxplot_response_time_00-06_25_dagger}
\end{subfigure}%
\begin{subfigure}{.5\textwidth}
  \centering
  \vspace{-0.2cm}
  \includegraphics[width=.91\linewidth]{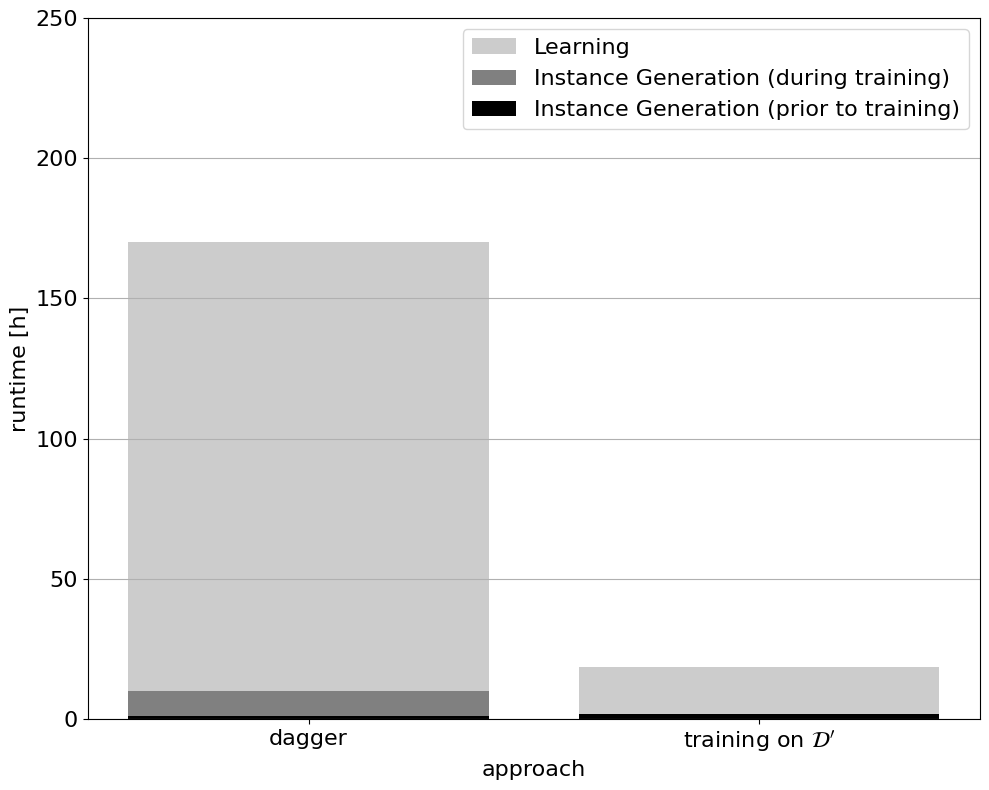}
  \caption{Runtimes: 25 Vehicles}
  \label{fig:dagger_runtime_comparison_25v}
\end{subfigure}

\begin{subfigure}{.5\textwidth}
  \centering
  \includegraphics[width=.95\linewidth]{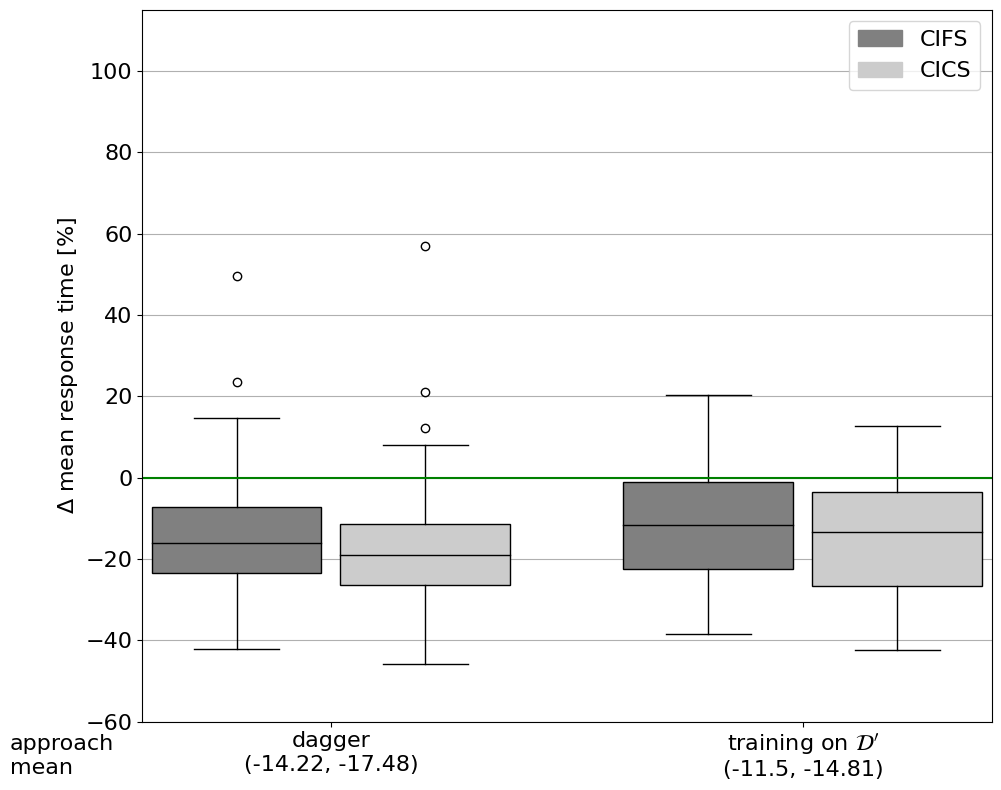}
  \caption{Mean Response Time Improvements: 50 Vehicles}
  \label{fig:boxplot_response_time_00-06_50_dagger}
\end{subfigure}%
\begin{subfigure}{.5\textwidth}
  \centering
  \vspace{-0.2cm}
  \includegraphics[width=.91\linewidth]{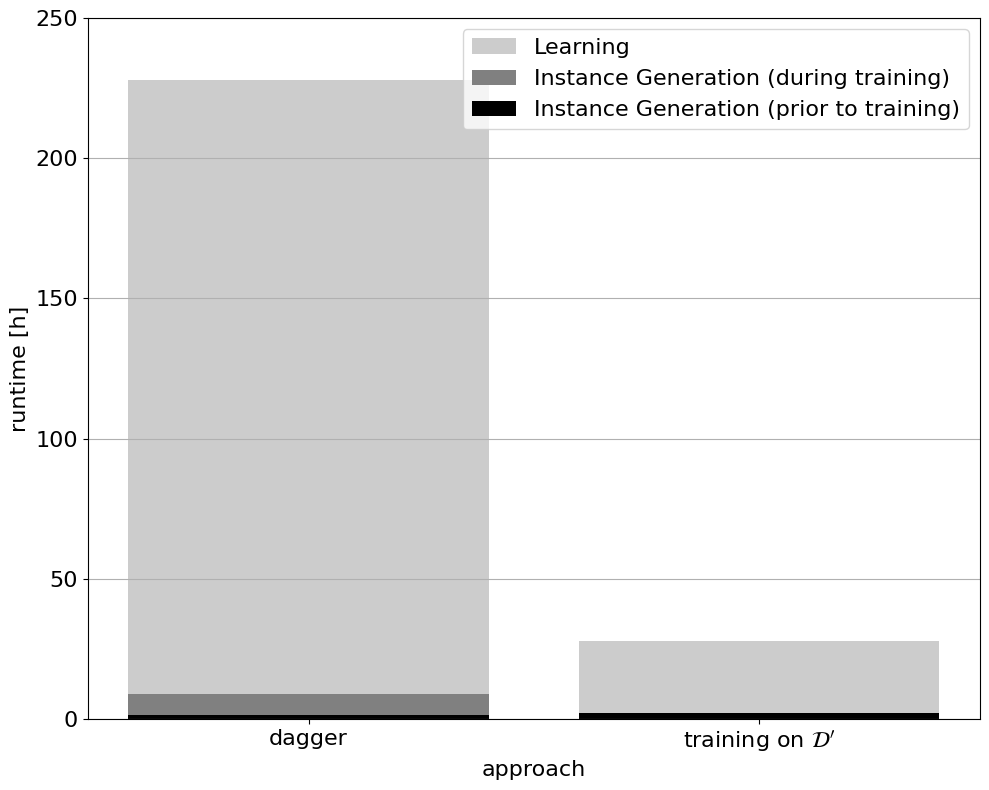}
  \caption{Runtimes: 50 Vehicles}
  \label{fig:dagger_runtime_comparison_50v}
\end{subfigure}

\caption{Approach comparison based on mean response time and runtime}
\label{fig:dagger_comparison}
\end{figure}

Training the predictor on the enhanced data set $\mathcal{D}'$ achieves significant runtime reductions. For the high-resource scenario, the total runtime reduces by 86.9\%, from 230 hours to 30 hours, while for the low-resource scenario, it reduces by 87.9\%, from 171 hours to 21 hours. Although \texttt{DA\scriptsize{GGER}} shows slightly better performances in terms of mean response time, the improvements are relatively small. The mean response time improvement can be increased by 0.91 percentage points in the low-resource scenario and by approximately 2.7 percentage points in the high-resource scenario. 
Despite these performance improvements, \texttt{DA\scriptsize{GGER}}'s iterative training is computationally intensive, which makes \texttt{DA\scriptsize{GGER}} impractical for large instances, e.g., for high-demand scenarios. In contrast, generating $\mathcal{D}'$ a priori is computationally efficient, offering substantial runtime savings while still delivering significant performance improvements over both online benchmarks (see Figures \ref{fig:boxplot_response_time_daily}-\ref{fig:dagger_comparison}). Even though the results achieved by \texttt{DA\scriptsize{GGER}} are superior, training the model on $\mathcal{D}'$ yields competitive results, demonstrating its practical advantages, particularly in scenarios with limited computational resources or when scalability is critical.

{\result{Training on the enhanced dataset $\mathcal{D}'$ offers substantial runtime savings of up to 87.9\% with competitive performance, making it a practical and scalable alternative to the computationally intensive \texttt{DA\scriptsize{GGER}} approach, especially in resource-constrained scenarios or when scalability is critical.}}
\newline
\section{Conclusion}\label{sec:conclusion}
We presented a CO-augmented ML pipeline for learning online policies to optimize ambulance dispatching and redeployment in EMS systems. As a basis, we first presented a MILP to generate optimal dispatching and redeployment decisions in offline settings. Second, we leveraged the derived optimal solutions to learn online policies in a supervised fashion. To improve the pipeline's performance, we enhanced the training set by incorporating states derived from following suboptimal policies, helping the model to learn more robust decision-making policies. Finally, we conducted a numerical case study on San Francisco's 911 call data to evaluate the performance of the learned policies, comparing them against two online benchmarks: closest idle dispatching and redeployment to either a fixed or the nearest station.

Our results demonstrate that the learned policies can outperform both online benchmarks across all tested scenarios, with a reduction in mean response time of up to 30\%. Analyzing the dispatching and redeployment decisions made by the different policies highlights the advantages of the learned policies, notably, their anticipatory decision-making strongly contributing to reduced response times. We further show that training dedicated weekday-models yields better results in scenarios with both the highest and lowest vehicle-request ratios compared to models trained on a mix of weekday and weekend data. Additionally, we show that augmenting the training set with states generated by following suboptimal policies prior to training can improve the learned policies' performances, particularly in resource-constrained scenarios where decision-making is more challenging. Compared to improving performance by utilizing the \texttt{DA\scriptsize{GGER}} approach during training, this approach offers substantial runtime savings of up to 87.9\% with competitive performance, making it a practical and scalable alternative. 

Future extensions of this work could explore the use of additional predictor models to be applied in the ML pipeline to further enhance the pipeline's performance. Additionally, focusing on the determination of promising waiting locations, given the learned dispatching and redeployment policy, presents a promising area for future research.


%
\singlespacing{
\bibliographystyle{model5-names}
\bibliography{mybibfile} 
\newpage
\onehalfspacing
\begin{appendices}
	\normalsize
	\section{Runtimes of Offline Solution Models: Soft vs. Hard Constraints}\label{sec:appendix_solution_model_with_soft_constraints}

The following shows an alternative model for the offline problem, ensuring that all feasible requests are served via soft constraints. For incentivizing request fulfillment via soft constraints, we introduce a sufficiently high penalty $\gamma$ for unserved requests. We adapt the objective (\ref{eq:MIP_objective}) to minimize the applied penalties and the sum of response times (\ref{eq:MIP_objective_soft}). We further split constraint (\ref{eq:request_assigned_to_max_1_path}) into two constraints to distinguish between optional and mandatory arcs that must be served: First, we allow emergency requests to not be served (\ref{eq:request_assigned_to_max_1_path_soft}). Second, we enforce that each vehicle serves exactly one path (\ref{eq:vehicles_serves_max_1_path_soft}). Constraints (\ref{eq:calculation_response_times}), (\ref{eq:dispatch_delays}), and (\ref{eq:direct dispatches})-(\ref{eq:variable_domain_alpha}) remain unchanged. 

\begin{equation}
    min \sum_{\substack{(i,j)\in {A}:\\j\in V^{\mathcal{\tilde{R}}}}}  -\gamma x_{ij} + \alpha_{ij}\label{eq:MIP_objective_soft}
\end{equation}
s.t.
\begin{align}
\sum_{i\in{V}} x_{ij} &\leq 1 &&\forall j \in {V^{\mathcal{\tilde{R}}}} \cup {V^{\mathcal{\hat{R}}}}: (i, j) \in {{A}} \label{eq:request_assigned_to_max_1_path_soft} \\
\sum_{i\in{V}} x_{ij} &= 1 &&\forall j \in {V^\mathcal{M}}: (i, j) \in {{A}} \label{eq:vehicles_serves_max_1_path_soft} \\
&(\ref{eq:calculation_response_times}), (\ref{eq:dispatch_delays}), (\ref{eq:direct dispatches}), (\ref{eq:flow_conservation}), (\ref{eq:variable_domain_x}), (\ref{eq:variable_domain_alpha})\nonumber
\end{align}

We compare both models' runtimes in Figure~\ref{fig:boxplot_runtime_mip}. 
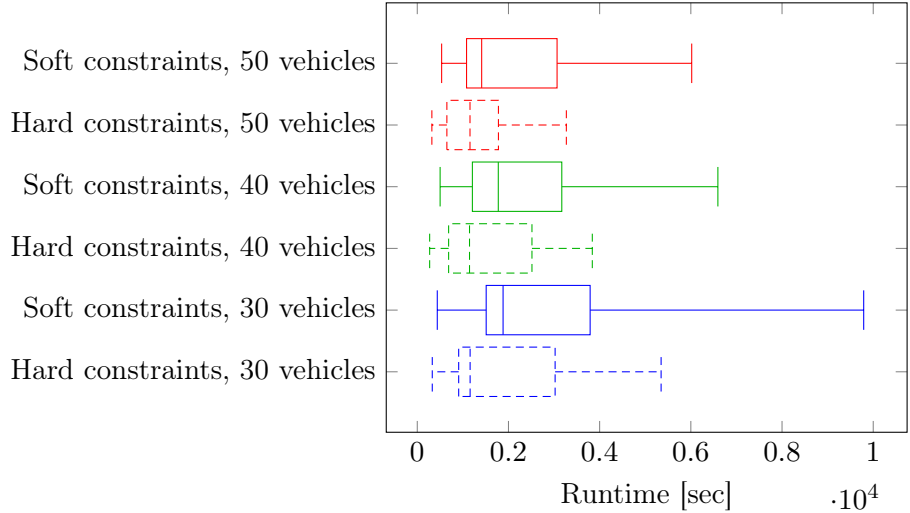
\begin{figure}[h]
  \centering
  \pgfplotsset{compat=1.8}
\usepgfplotslibrary{statistics}

\begin{tikzpicture}
  \begin{axis}
    [
    ytick={1,2, 3, 4, 5, 6},
    yticklabels={
        {Hard constraints, 30 vehicles}, 
        {Soft constraints, 30 vehicles},
        {Hard constraints, 40 vehicles},
        {Soft constraints, 40 vehicles},
        {Hard constraints, 50 vehicles},
        {Soft constraints, 50 vehicles}},
    xlabel={Runtime [sec]},
    ]
    \addplot+[
    boxplot prepared={
      lower whisker=329.0436429977417,lower quartile=908.505383014679,median=1159.054574966431,upper quartile=3023.3370459079742,upper whisker=5347.841787099838
    },
    densely dashed,
    color=blue,
    ] coordinates {};
    \addplot+[
    boxplot prepared={
      lower whisker=438.8322010040283,lower quartile=1511.458374619484,median=1882.026897907257,upper quartile=3791.350968003273,upper whisker=9790.989120960236
    },
    solid,
    color=blue,
    ] coordinates {};
    \addplot+[
    boxplot prepared={
      lower whisker=271.7959089279175,lower quartile=687.7542021274567,median=1146.1295759677887,upper quartile=2515.4363659620285,upper whisker=3838.928108215332
    },
    densely dashed,
    color=green!70!black,
    ] coordinates {};
    \addplot+[
    boxplot prepared={
      lower whisker=501.1383020877838,lower quartile=1208.7521179914474,median=1777.0080428123474,upper quartile=3169.770637989044,upper whisker=6593.099925041199
    },
    solid,
    color=green!70!black,
    ] coordinates {};
    \addplot+[
    boxplot prepared={
      lower whisker=320.28767490386963,lower quartile=649.3413000106812,median=1156.6553750038147,upper quartile=1780.08278298378,upper whisker=3269.7040231227875
    },
    densely dashed,
    color=red,
    ] coordinates {};
    \addplot+[
    boxplot prepared={
      lower whisker=534.6246619224548,lower quartile=1081.1118685007095,median=1412.6216928958893,upper quartile=3066.1873955726624,upper whisker=6019.123133897781
    },
    solid,
    color=red,
    ] coordinates {};
  \end{axis}
\end{tikzpicture}
  \caption{Runtime Comparison: MILP with Hard Constraints vs. with Soft Constraints}
  \label{fig:boxplot_runtime_mip}
\end{figure}
Both models lead to identical dispatching and redeployment decisions as all requests can be served, and we choose a sufficiently high penalty $\gamma$ in objective (\ref{eq:MIP_objective_soft}) to ensure their fulfillment. For this analysis, we generated optimal solutions with both models operating 50, 40, and 30 ambulances for the period from 12 p.m. to 6 p.m. throughout October 2023. Our findings show that using hard constraints is superior in terms of model runtimes. Thus, provided that the graph structure allows all requests to be served, hard constraints are runtime-wise more efficient for guaranteeing request fulfillment. Nevertheless, for graph structures for which requests may become unfeasible, e.g., when implementing maximum driving times for ambulances, applying soft constraints to model request fulfillment constitutes a viable alternative.

\section{Maximum Covering Location Problem for Ambulance Allocation}\label{sec:appendix_coverage_model}
To fairly allocate ambulances $m \in \mathcal{M}$ to stations $s \in \mathcal{S}$ that are used as their home bases, we apply the MCLP, originally introduced by \cite{church1974maximal}. In this context, we divide the examined area into a set of geographical regions $g \in \mathcal{G}$ in which a percentage share of $0 \leq \mu_g \leq 1$ emergencies arise, where $\sum_{g \in \mathcal{G}} \mu_g = 1$. A region is covered if it can be reached within a maximum driving time $\kappa$. We denote the set of stations covering region $g$ by  $\mathcal{S}^C_g = \{s \in \mathcal{S}: \tau({l_s,c_g})\leq \kappa\}$ where $c_g$ is the center of $g$.  We extend the original model by assuming that each station can host up to $a$ ambulances and that the minimum number of ambulances that must cover a region is $\zeta$. We introduce the binary variables $x_s$ and $y_g$, denoting the number of ambulances allocated to station $s$ and the number of ambulances covering region $g$, correspondingly. In this setting, the following model maximizes the coverage of emergency incidents occurring in the geographical regions.
\begin{equation}
        max \sum_{g \in {\mathcal{G}}}  \mu_gy_g \label{eq:MIP_coverage_objective}
\end{equation}
s.t.
\begin{align}
      \sum_{s \in \mathcal{S}^C_g} x_s &\geq y_g \ &\forall g \in {\mathcal{G}}\label{eq:coverage_constraint}\\
      y_g &\geq \zeta \ &\forall g \in {\mathcal{G}}\label{eq:min_coverage_constraint}\\   
      \sum_{s \in \mathcal{S}} x_s &= |\mathcal{M}|\label{eq:max_nr_ambulances}\\
      x_s &\leq a \ &\forall s \in \mathcal{S} \label{eq:max_nr_ambulances_per_station} \\
      x_s &\in \mathbb{N}_{\geq 0} \ &\forall s \in \mathcal{S}  \label{eq:variable_domain_x_s}\\   
      y_g &\in \mathbb{N}_{\geq 0} \ &\forall g \in \mathcal{G}   \label{eq:variable_domain_y_g}
\end{align}
Objective (\ref{eq:MIP_coverage_objective}) maximizes the coverage of the geographical regions weighted by their percentage share of emergency calls. Constraint ({\ref{eq:coverage_constraint}}) and ({\ref{eq:min_coverage_constraint}}) ensure that each region is covered by at least $\zeta$ ambulances. Constraint (\ref{eq:max_nr_ambulances}) and (\ref{eq:max_nr_ambulances_per_station}) limit the total number of
ambulances and the number of ambulances allocated to each station, correspondingly. Constraints (\ref{eq:variable_domain_x_s}) and (\ref{eq:variable_domain_y_g}) define the variable domains. Applying the MEXCLP, a commonly applied probabilistic model accounting for vehicle availability originally introduced by \cite{daskin1983maximum}, yields similar ambulance allocations. For the sake of simplicity, we apply the simpler MCLP model.

For the numerical case study presented in Section \ref{sec:castestudy}, we solve this model based on the ambulance demand from January to June 2023. As a basis, we divide the region into hexagons – each with an area of $0.737\text{km}^2$ applying the hexagonal hierarchical geospatial indexing system H3 \citep{brodsky2018}. We enforce double coverage ($\zeta=2$) and apply a commonly applied coverage threshold $\kappa$ of 10 minutes \citep{aboueljinane2013review}, i.e., each region is covered by at least two ambulances that can reach the region's center within 10 minutes. We further allow at most two ambulances per station ($a=2$). The resulting allocation of ambulances provides the basis for the static redeployment policy, in which each ambulance always returns to its assigned station. The allocation is further used to initialize the system for the remaining policies.

\section{Kolmogorow-Smirnow Tests for Emergency Call Patterns}\label{sec:KS_test}
To evaluate the similarity of spatial emergency call distributions across different days, we conduct Kolmogorow-Smirnow (KS) tests. These allow us to test the null hypothesis, stating that two samples were drawn from the same distribution. Applying a confidence level of 99\%, we reject the null hypothesis if the p-value is $<0.01$. As a basis, we divide the region into hexagons – each with an area of $0.737\text{km}^2$ applying the hexagonal hierarchical geospatial indexing system H3 \citep{brodsky2018}. We calculate the average number of emergency calls per cell and day for two time periods: the low-demand period (12 a.m. to 6 a.m.) and the high-demand period (12 p.m. to 6 p.m.). We show the obtained p-values for both time ranges in Figure~\ref{fig:ks_tests}. 
\begin{figure}[h!]
\centering
\begin{subfigure}{.5\textwidth}
  \centering
  \includegraphics[width=.95\linewidth]{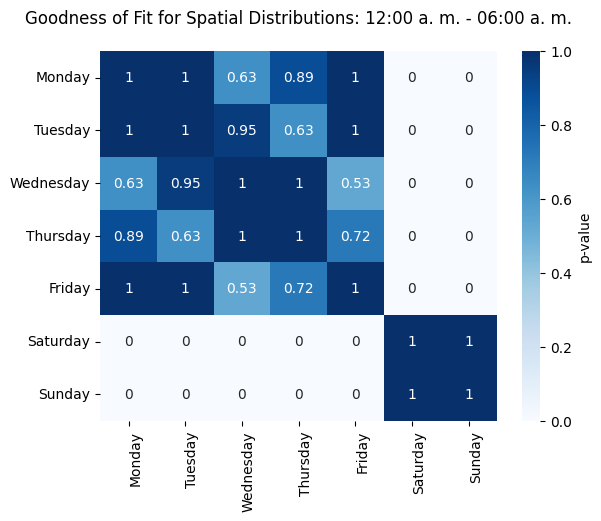}
  \label{fig:sub1}
\end{subfigure}%
\begin{subfigure}{.5\textwidth}
  \centering
  \includegraphics[width=.95\linewidth]{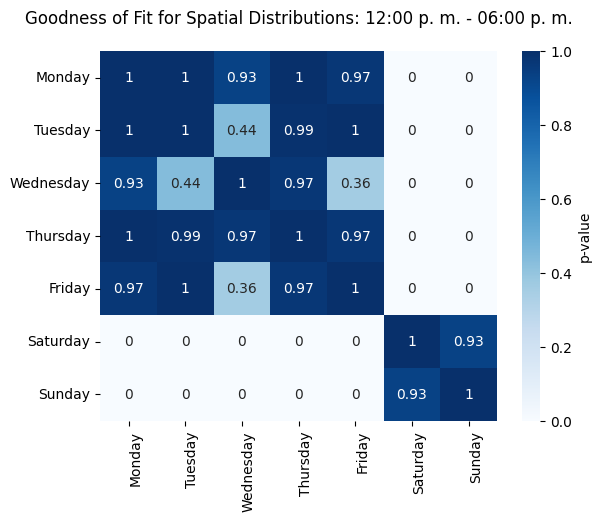}
  \label{fig:sub2}
\end{subfigure}
\caption{Kolmogorow-Smirnow Tests}
\label{fig:ks_tests}
\end{figure}
We see that for both time periods the spatial emergency call distributions are relatively consistent across weekdays (Monday - Friday). For all weekday combinations, the KS tests indicate that the samples were drawn from the same distribution. Similarly, the distributions of Saturdays and Sundays are consistent for both time periods.

\end{appendices}
\end{document}